\documentclass[journal]{IEEEtran}
\usepackage{amsmath,amsfonts}
\usepackage{array}
\usepackage[caption=false,font=normalsize,labelfont=sf,textfont=sf]{subfig}
\usepackage{textcomp}
\usepackage{stfloats}
\usepackage{url}
\usepackage{verbatim}
\usepackage{graphicx}
\usepackage{xcolor}     
\usepackage{color}
\usepackage{colortbl}
\usepackage{cite}
\usepackage{xspace}
\usepackage{bbm}
\usepackage{amsmath}
\usepackage{enumitem}
\usepackage{caption}
\usepackage[colorlinks=true,linkcolor=blue,breaklinks=true,anchorcolor=blue,citecolor=blue]{hyperref}
\usepackage{cleveref}

\usepackage{algorithmic}
\usepackage{algorithm}

\renewcommand{\emph}{\textit}
\newcommand{\M}{MENTOR\xspace}
\newcommand{\DDC}{Dynamic Distance Constraint}
\newcommand{\EED}{Exploration-Exploitation Decoupling}
\newcommand{\hi}{\mathrm{hi}}
\newcommand{\lo}{\mathrm{lo}}
\newcommand{\hf}{\text{hf}}
\newcommand{\rnd}{\text{rnd}}
\newcommand{\g}{\text{g}}
\newcommand{\sub}{\text{sub}}

\begin{document}
\ifdefined\includedbyb
\else
\fi
\title{MENTOR: Guiding Hierarchical Reinforcement Learning with Human Feedback and Dynamic Distance Constraint}

\author{
Xinglin~Zhou$^*$,~\IEEEmembership{}
Yifu~Yuan$^*$,~\IEEEmembership{} 
Shaofu~Yang,~\IEEEmembership{Member,~IEEE,} 
and~Jianye~Hao,~\IEEEmembership{Senior Member,~IEEE}

\thanks{This work was supported by Southeast University Big Data Computing Center. Corresponding author: Shaofu Yang.}

\thanks{X. Zhou and Y. Yuan contributed equally to this work and are considered co-first authors.}

\thanks{This work was supported by Southeast University Big Data Computing Center. Corresponding author: Shaofu Yang.}

\thanks{S. Yang is with the School of Computer Science and
 Engineering, Southeast University, Nanjing 210096, China (e-mail: sfyang@
 seu.edu.cn).}

\thanks{X. Zhou is with the Southeast University-Monash University Joint Graduate School, Southeast University, Suzhou 215123, China (e-mail: 220214855@seu.edu.cn).}
 
\thanks{J. Hao and Y. Yuan are with the College of Intelligence and Computing, Tianjin University, Tianjin 300072, China (e-mail: jianye.hao@tju.edu.cn; yuanyf@tju.edu.cn).}
}


\markboth{IEEE TRANSACTIONS ON EMERGING TOPICS IN COMPUTATIONAL INTELLIGENCE}{}

\maketitle

\begin{abstract}
Hierarchical reinforcement learning (HRL) provides a promising solution for complex tasks with sparse rewards of agents, which uses a hierarchical framework that divides tasks into subgoals and completes them sequentially. However, current methods struggle to find suitable subgoals for ensuring a stable learning process. To address the issue, we propose a general hierarchical reinforcement learning framework incorporating hu\underline{m}an f\underline{e}edback and dy\underline{n}amic dis\underline{t}ance c\underline{o}nst\underline{r}aints, termed \textbf{\M}, which acts as a ``\textit{mentor}''. Specifically, human feedback is incorporated into high-level policy learning to find better subgoals. Furthermore, we propose the Dynamic Distance Constraint~(DDC) mechanism dynamically adjusting the space of optional subgoals, such that \M can generate subgoals matching the low-level policy learning process from easy to hard. As a result, the learning efficiency can be improved. As for low-level policy, a dual policy is designed for exploration-exploitation decoupling to stabilize the training process.  Extensive experiments demonstrate that \M uses a small amount of human feedback to achieve significant improvement in complex tasks with sparse rewards. Further details and code implementations can be found at \href{https://github.com/nidesuipao/MENTOR.git}{https://github.com/nidesuipao/MENTOR.git}.

\end{abstract}


\section{Introduction}

\IEEEPARstart{T}{he} problem of sparse reward is consistently challenging in the domain of reinforcement learning (RL) \cite{reddy2019sqil, vecerik2017leveraging, luo2023rarsmsdou}, attributing to two main factors: challenging exploration, and unstable training. In recent years, several approaches have been proposed to relieve these issues, including goal-conditional reinforcement learning \cite{andrychowicz2017hindsight}, curiosity-driven exploration \cite{burda2018exploration, gao2023dynamic} and hierarchical reinforcement learning (HRL) \cite{levy2017learning, nachum2018data, hutsebaut2022hierarchical, 10246843}.

Hierarchical Reinforcement Learning (HRL) is effective for long-horizon tasks with sparse rewards, as it decomposes tasks into more manageable subgoals, mitigating challenges related to exploration and unstable training. However, there are two primary challenges in its practical applications.
\emph{1) Generating effective subgoals.} To create efficient subgoals that guide low-level policies, manual design \cite{dietterich2000hierarchical, parr1997reinforcement} and automatic generation methods \cite{zhang2022adjacency, eysenbach2018diversity, park2023controllability, hutsebaut2022hierarchical} have been proposed. However, manual design is resource-intensive and struggles with complex tasks \cite{ahn2022can}, while automatic generation demands significant computational resources to explore the entire state space \cite{campos2020explore}.
\emph{2) Efficient subgoal completion.} During low-level learning, hindsight relabeling \cite{andrychowicz2017hindsight} adjusts subgoals to turn failed transitions into successes but lacks the capacity for effective exploration. Curiosity-driven techniques, such as Random Network Distillation (RND), prevent revisiting states \cite{bougie2022hierarchical} but can destabilize training due to reward bonuses associated with exploration, particularly in sparse reward settings. Frequent failures in low-level subgoal achievement can lead to non-stationarity at the high level. While some studies \cite{levy2017learning, nachum2018data} address these issues through hindsight mechanisms, e.g., penalizing high-level policies \cite{levy2017learning}, they do not fully resolve the problem of ensuring that low-level policies consistently accomplish tasks.

\subsection{Related work}

\subsubsection{Hierarchical reinforcement learning} In the field of HRL, identifying meaningful subgoals within long-horizon tasks has been extensive research. This includes studies on options \cite{bacon2017option, harb2018waiting, bagaria2019option}, goals \cite{andrychowicz2017hindsight, levy2017learning, schaul2015universal} and skills \cite{eysenbach2018diversity, park2023controllability, sharma2019dynamics, levy2023hierarchical}. Manual-designed subgoals are costly and challenging for complex tasks \cite{ahn2022can}. Automatic learning of meaningful subgoals without any guidance from an external expert is a significant challenge \cite{jinnai2019finding}. CSD \cite{park2023controllability} aims to discover latent skills via mutual-information maximization. However, combining meaningful skills into task completion is a continuous challenge. Director \cite{hafner2022deep} introduces a practical method for learning hierarchical behaviors directly from pixels within a learned world model. \textcolor{black}{\cite{WangWHLL23} proposes a skill-based hierarchical reinforcement learning (SHRL) framework for solving the problem of visual navigation of a target. The SHRL framework consists of a high-level policy and three low-level skills: search, adjustment, and exploration. \cite{KimSS21} introduces a novel framework called HIGL, which effectively reduces the action space of high-level policies by sampling informative landmarks for exploration and training the high-level policy to generate subgoals towards selected landmarks. HAC \cite{levy2017learning} addresses the challenges of sparse reward and non-stationarity at high-levels through the utilization of hindsight action transitions and hindsight goal transitions. HIRO \cite{nachum2018data} employs a model to generate a new high-level action to rectify the high-level transitions. AGILE \cite{wang2022hierarchical} addresses the non-stationarity issue by using adversarial learning to guide the generation of compatible subgoals for the low-level policy. Previous work has centered on resolving non-stationary issues or on task decomposition strategies, yet often faces the challenge of sparse rewards, prompting unnecessary exploration. Our study introduces human guidance and DDC to decompose tasks effectively. Human guidance directs subgoals, reducing reward sparsity, while DDC controls subgoal difficulty, synchronizes learning across levels, and relieves non-stationarity issues.}


\subsubsection{Reinforcement Learning from Human Feedback}
The surge in popularity of ChatGPT \cite{brown2020language, ouyang2022training} has significantly boosted the recognition of RLHF in recent times. RLHF is a technique for aligning the functionalities of Large Language Models (LLMs) with human ethical considerations and preference, achieved by integrating human feedback into the learning process \cite{lee2021pebble, lee2021b, park2022surf, RafailovSMMEF23, DaiPSJXL0024}. For instance, \cite{wu2023better} demonstrates how human preferences can be integrated into text-to-image generation models, aligning them with human aesthetics. \cite{zhan2021human} introduces a GAN-augmented reinforcement learning strategy that efficiently learns robot behaviors through human preferences, significantly reducing the reliance on human demonstrations. RLHF learns reward functions from pairwise comparison and ranking based on human preference \cite{ouyang2022training, lee2021pebble, shah2016estimation, saha2022efficient}. Human intuition and experience can be incorporated as guidance in high-level decision-making, particularly in setting subgoals within the HRL framework \cite{bougie2022hierarchical}. Nevertheless, humans may struggle to offer immediate guidance that corresponds with the agent's capabilities.

\subsection{Our Contribution}
We introduce a novel framework, MENTOR, which integrates human guidance into the high-level policy learning process. This is achieved through RLHF, a method that trains a reward model by learning human preferences via binary comparisons of subgoals. MENTOR utilizes this reward model to generate subgoals at the high-level, effectively steering the agent towards optimal behavior. Additionally, we introduce DDC. It measures subgoal difficulty by distance and adjusts the subgoal space accordingly. To enable an agent to quickly complete subgoals at the low-level, we introduce {\EED} (EED), which uses one policy to explore while the other policy learns from the experience of the exploring policy to stabilize the training. We summarize the main contributions as follows:
\begin{itemize}
\item We propose {\M}, leveraging human feedback to guide the subgoal direction and {\EED} to simultaneously realize exploration and exploitation in subgoal attainment.
\item We introduce {\DDC}, dynamic aligning the subgoal difficulty to the capabilities of the low-level policy.
\item We demonstrate that {\M} outperforms other baselines in accomplishing tasks with sparse rewards across various domains.
\end{itemize}

\section{Preliminary}

\subsection{Problem Setting}
Define Markov Decision Process with a set of goals, characterized by the tuple $\langle {\mathcal{S}},{\mathcal{G}}, {\mathcal{A}},{P},{r},\gamma \rangle$, where $\mathcal{S}$, $\mathcal{G}$, and $\mathcal{A}$ are the sets of state, goal, and action, respectively, ${P(s^{{\prime}}|s,a)}$ is the transition probability function, $r_{\g}(s, g): \mathcal{S}\times \mathcal{G} \to \mathbb{R}$ is the reward function, and  $\gamma$ is the discount rate $\in [0, 1)$. The objective is to find an optimal policy $\pi^*$ such that
$$\pi^* = \arg\max_{\pi} \mathbb{E}_{\mathcal{T} \sim \pi} \left[\sum_{t=1}^{H}\gamma^t r_{\g}(s_t,g_t)\right],$$
where $\mathcal{T}$ denotes a trajectory generated under the policy $\pi$ starting from an initial state. The reward function can be defined as $r_{\g}(s, g)=\mathbbm{1} (\Vert g-s \Vert_2<\epsilon)$, which $\epsilon$ is a distance threshold determining contact.

\subsection{Hindsight Relabelling}
Hindsight relabelling \cite{andrychowicz2017hindsight} addresses sparse reward in GCRL by redefining failed transitions as successful transitions, thus generating positive rewards. Specifically, for a failed transition $(s_t, a_t, r_t, s_{t+1}, g)$ with $r_t=r_{\g}(s_{t+1}, g) = 0$, it resets the transition as $(s_t, a_t, 1, s_{t+1}, g^{\prime})$ with goal $g^{\prime} = s_{t+1}$. By utilizing this new transition, it becomes possible to train an off-policy RL algorithm with more positive rewards. \cite{andrychowicz2017hindsight} also delve three heuristic methods for goal relabeling to enhance learning efficiency: (1) Future Sampling, which selects goals from subsequent states within the same trajectory; (2) Episode-Specific Random Sampling, which randomly selects goals from the same episode without considering their order; (3) Random Sampling, which chooses new goals from the entire dataset. In our implementation, we have opted for the Future Sampling strategy.

\subsection{Curiosity-driven Exploration}
Intrinsic motivation has been utilized to encourage agents to learn about their surroundings, even when extrinsic rewards are scarce \cite{burda2018exploration, bellemare2016unifying, pathak2017curiosity}. Curiosity-driven exploration is an intrinsic motivation strategy, exemplified by algorithms like RND, which fosters learning through the agent's desire to discover new information. In this framework, RND \cite{burda2018exploration} serves as the intrinsic reward. RND consists of two neural networks, represented as $f: \mathcal{S} \rightarrow {\mathbb{R}^k}$ and $\hat{f}: \mathcal{S} \rightarrow {\mathbb{R}^k}$, which are both randomly initialized and capable of transforming observations into embeddings. By fixing one network and training the other to minimize the prediction error, we follow the distillation optimization process:

\begin{equation}
    \begin{aligned}
        \mathbb{E}_{s \sim \rho_{\pi(s)}} \min_{\hat{f}} \Vert \hat{f}(s) - f(s) \Vert^2,
    \end{aligned}
    \label{distillation loss}
\end{equation}
where $s$ is sampled from the replay buffer. Upon the agent's interaction with the environment, yielding the current state $ s_t $, we proceed to compute RND reward:
\begin{equation}
    \begin{aligned}
        r_{\rnd}(s_{t}) = \Vert \hat{f}(s_{t}) - f(s_{t}) \Vert^2.
    \end{aligned}
    \label{rnd reward}
\end{equation}

This RND reward encourages the agent to visit novel states, as it will be higher in regions of the state space that the predictor network finds difficult to approximate, thus fostering exploration in the learning process. To ensure effective policy optimization and stability in reinforcement learning, it is essential that the RND reward, as an intrinsic reward, is normalized to align with the scale of extrinsic rewards:

\begin{equation}
    \begin{aligned}
        r_{\text{intri}}(s_{t}) = \frac{r_{\rnd}(s_{t}) -\mu(r_{\rnd})}{\sigma(r_{\rnd})}.
    \end{aligned}
    \label{intrinsic reward}
\end{equation}

We determine the normalized intrinsic reward for state $ s_{t} $ by converting the RND reward into a Z-score, aligning it with the mean $ \mu(r_{\rnd}) $ and scaling it according to the standard deviation $ \sigma(r_{\rnd}) $.

\section{Methodology}

\begin{figure*}[tp!] 
  \centering
  \includegraphics[width=\textwidth]{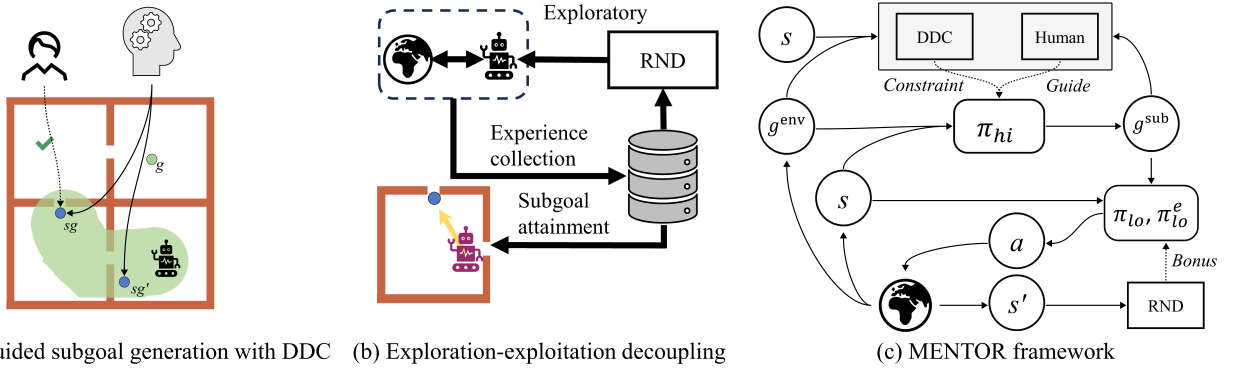} 
  \caption{(a) The high-level policy selects subgoals with DDC (shades of green), and human guides by comparing these subgoals. (b) The low-level decouples exploration and exploitation through two policies, one policy explores the environment and the other learns from the experience of exploration. (c) Diagrammatic representation of {\M} framework.}
  \label{guru}
\end{figure*}

\textcolor{black}{HRL consists of multiple levels, each focused on solving specific problems. The high-level is responsible for providing overall guidance, while the low-level handles task execution. This difference in responsibilities between the levels requires distinct reward function characteristics. Depending on unique features of diverse levels, we propose \M, shown in \Cref{guru}(c). At the high-level, it utilizes RLHF and DDC to address the challenge of generating instructive subgoals shown in \Cref{guru}(a). At the low-level, it decouples exploration and exploitation, solving the instability associated with curiosity-driven exploration shown in \Cref{guru}(b).}

Before introducing the framework, we briefly review here to introduce notation. The framework consists of four neural networks: high-level policy $\pi_{\hi}: \mathcal{S} \times \mathcal{G} \rightarrow \mathcal{G}$, low-level policy $\pi_{\lo}: \mathcal{S} \times \mathcal{G} \rightarrow \mathcal{A}$, reward model learned by human feedback $r_{\hf}: \mathcal{S} \times \mathcal{G} \times \mathcal{G} \rightarrow [0, 1]$, RND model $f_{\rnd}: \mathcal{S} \rightarrow [0, 1]$ and distance model $d: \mathcal{G} \times \mathcal{G} \rightarrow [0, 1]$ with parameters $\theta_{\hi}$, $\theta_{\lo}$ , $\theta_{\hf}$, $\theta_{\rnd}$ and $\theta_{d}$ respectively. The distance model, denoted as $ d $, operates within the Dynamic Distance Constraint. In a given episode, $\pi_{\hi}$ initially proposes subgoal $g^{\sub}_0$ based on current state $s$ and environment goal $g^{\text{env}}$, followed by task given by $\pi_{\lo}$ to achieve $g^{\sub}_0$. If $\pi_{\lo}$ succeeds and the episode remains active, $\pi_{\hi}$ subsequently issues a new subgoal $g^{\sub}_1$. Thus, the high-level trajectory is represented as ($s_{t_0}, s_{t_1},...,s_{t_{n-1}}$), where ${t_i}$ signifies the moment of the $i$-th subgoal generation by $\pi_{\hi}$. Concurrently, the low-level trajectory unfolds as ($s_{0}, s_{1},...,s_{L-1}$), $L$ denotes the trajectory length. 

\subsection{\textcolor{black}{RLHF and {\DDC} in High-level}} 
\subsubsection{\textcolor{black}{Subgoal generation using RLHF}}
Subgoal generation poses a significant challenge, with manual setup being costly and difficult, while automatic methods demand extensive computational resources to explore the state space. High-level subgoal generation requires macro, abstract, and generalized guidance, closely related to human preferences. RLHF offers human preferences through sample comparisons. \textcolor{black}{Despite the inherent uncertainty and noise in the human feedback acquired through pairwise preference comparisons, this approach for reward model learning effectively guides the high-level policy learning because HRL mitigates the issue of uncertainty and noisy preferences in RLHF, creating a mutual benefit for both HRL and RLHF.} 

RLHF uses pairwise comparison to train a reward model which can be used as the reward function for the high-level policy. The training process consists of (1) extracting pairs $(s_1, g^{\sub}_1, g^{\text{env}})$ and $(s_2, g^{\sub}_2, g^{\text{env}})$ from the high-level replay buffer. (2) Human annotators provide pairwise feedback $v$ which 0 prefers $(s_1, g^{\sub}_1, g^{\text{env}})$ , $1$ prefers $(s_2, g^{\sub}_2, g^{\text{env}})$ and $0.5$ for same preference. Preference can be assessed by the distance to the environment goal $g^{\text{env}}$. (3) After collecting the feedback dataset into the reward model buffer $B_{\hf}$, we can train the reward model $r_{\hf}$ by optimizing a modified Bradley-Terry objective \cite{bradley1952rank}. We define the possibility that human prefers $(s_1, g^{\sub}_1, g^{\text{env}})$ to $(s_2, g^{\sub}_2, g^{\text{env}})$:
\begin{equation}
\begin{aligned}
&P \left [(s_1, g^{\sub}_1) \succ (s_2, g^{\sub}_2)| g^{\text{env}}\right ] \\
&\quad = \frac{\exp(r_{\hf}(s_1,g^{\sub}_1,g^{\text{env}}))}{\sum_{i=1}^2 \exp(r_{\hf}(s_i,g^{\sub}_i,g^{\text{env}}))}.
\end{aligned}
\end{equation}
Then, we define two probabilities:
$$
p_1 = P[(s_1, g^{\sub}_1) \succ (s_2, g^{\sub}_2) | g^{\text{env}}],
$$
$$p_2 = P[(s_2, g^{\sub}_2) \succ (s_1, g^{\sub}_1) | g^{\text{env}}], 
$$
and the modified Bradley-Terry objective is as follows:

\begin{equation}
\begin{aligned}
    \max_{r_{\hf}} \mathbb{E}_{(s_1, g^{\sub}_1,s_2,g^{\sub}_2, g^{\text{env}}, v)\sim B_{\hf}} \left[(1-v) \log (p_1) + v\log (p_2) \right].
\end{aligned}
\label{reward update}
\end{equation}

In optimizing the high-level policy, the high-level reward function is set to be $r_{\hi} = r_{\hf}(s, g^{\sub}, g^{\text{env}})$. In the dynamic process of execution, humans and algorithms engage in ongoing interactions, where high-level policies can be subject to real-time guidance from human operators.

\subsubsection{\textcolor{black}{{\DDC} for Subgoal Difficulty Adjustment}}
while humans can provide direction for high-level goals by expressing preferences, they may struggle to define subgoals that align with the low-level policy's capabilities. It is quite possible that influenced by their cognition, humans might prefer subgoals that are close to the ultimate goal. Consequently, if the reward model learned by human preferences is designed to assign the highest rewards when the subgoal coincides with the goal, it could lead to a scenario where the high-level policy, without considering the low-level's capacity and basing its decisions solely on human feedback, generates subgoals that rapidly converge towards the goal. This approach could render the subgoals excessively challenging and diminish their instructional value. Thus multi-level simultaneous learning may be uncoordinated. 

\begin{figure}[tp!]
    \centering
    \includegraphics[width=\linewidth]{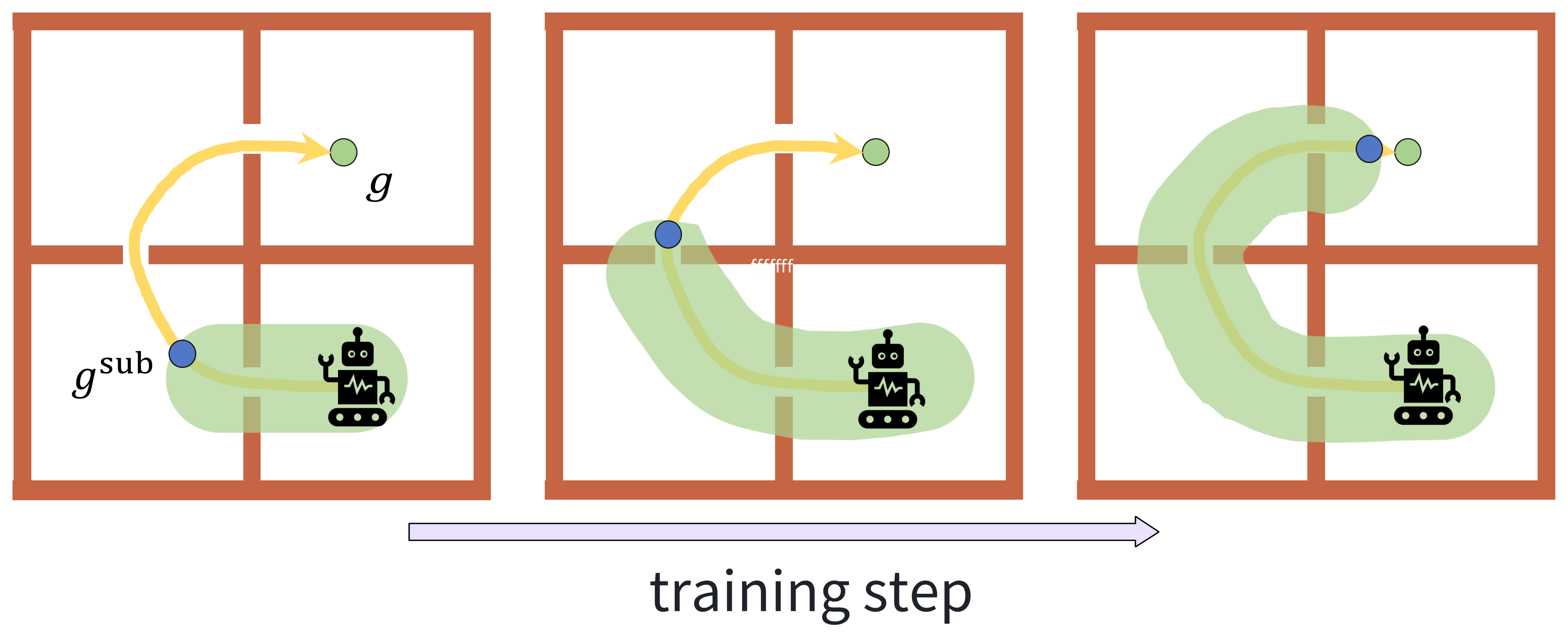}
    \caption{As the low-level capability improves, the DDC progressively relaxes, allowing the high-level to propose increasingly challenging subgoals.}
    \label{DDC}
\end{figure}

To solve this issue above, we introduce DDC, whose function is illustrated in \Cref{DDC}. This method limits the range of subgoals based on their distance from the current state (the variation in green shading depicted in the figure). \textcolor{black}{Under the function of DDC, the learning process of our framework {\M} is similar to curriculum learning\cite{zhang2020automatic, klink2024benefit, tzannetos2024proximal}. Curriculum learning facilitates a sequential learning process where simpler subgoals are mastered before progressing to more complex ones, thus laying a solid foundation for advanced subgoal acquisition.} This is achieved through a specific formulation:
\begin{equation}
    \begin{aligned}
        \max_{\pi_{\hi}} &\ \mathbb{E}_{g^{\sub}\sim \pi_{\hi}} \left[\sum_{i=0}^{n} r_{\hi} (s_{t_i}, g^{\sub}_i, g^{\text{env}}) \right]\\
        \mbox{s.t. } & \mathcal{H}(g^{\sub}_i, g^{\text{ach}}_{t_i}, k):= \max\{d(g^{\sub}_i, g^{\text{ach}}_{t_i})-k, 0\} \leq 0, 
    \end{aligned}
    \label{constraint high policy update}
\end{equation}      
where $d(g^{\sub}_i, g^{\text{ach}}_{t_i}): \mathcal{G} \times \mathcal{G} \rightarrow [0, 1]$ represents the distance between the subgoal $g^{\sub}_i$ and the achieved goal $g^{\text{ach}}_{t_i}$. The parameter $k$ sets the subgoal space range. The constraint ensures that the distance between the high-level subgoal and the current achieved goal remains within a range of $ k $ lengths, and we can adjust $ k $ to progressively reduce the difficulty. However, in scenarios like the Four Rooms domain, the Euclidean distance as $d(\cdot)$ may not accurately assess the difficulty of completing a task. For instance, a goal in the top right might be hard to reach despite a low Euclidean distance, indicating the need for a learned step distance measure. DDL \cite{hartikainen2019dynamical} offers a methodology for training distance models by randomly sampling state pairs from trajectories, to approximate the number of steps between states. Nevertheless, it's crucial to evaluate the step distance between unreachable subgoals and the current state. If the low-level policy fails to reach a subgoal, assigning a high distance value to these unreached subgoals relative to the current state is essential. Neglecting to do so, high-level policies might incorrectly view challenging subgoals as easy, leading to unrealistic subgoal proposals. To tackle this issue, we recommend incorporating extra samples containing unreached subgoals into the distance model objective:

\begin{equation}
    \begin{aligned}
        &\min_{d} \frac{1}{2} \mathbb{E}_{\tau \sim \rho_{\pi_{\lo}},i\in[0, L],j\in[i,L], g^{\sub} \sim \pi_{\hi}(\cdot|s_{0}, g^{\text{env}})} \\
    &\left[(d(s_i, s_j)-\frac{|i-j|}{L})^2 + \mathbbm{1}(s_L \neq g^{\sub})(d(s_i, g^{\sub}) - 1)^2 \right]. 
    \end{aligned}
    \label{distance update}
\end{equation}

The objective is formulated by minimizing the expected loss across trajectories $\tau$, sampled using the low-level policy $\rho_{\pi_{\lo}}$ from recent episodes, and the goals $g^{\text{env}}$ drawn from the environment's distribution $\rho_{g^{\text{env}}}$. $L$ is the length of the trajectory.

\textcolor{black}{In \Cref{constraint high policy update}, optimizing is challenging because of the strict constraint. To overcome this difficulty, we utilize the penalty function method \cite{LinMX22, bertsekas1997nonlinear}, which allows us to establish an unconstrained optimization objective:}

\begin{equation}
    \max_{\pi_{\hi}} \mathbb{E}_{g^{\sub}\sim \pi_{\hi}} \sum_{i=0}^{n}[r_{\hi}(s_{t_i}, g^{\sub}_i, g^{\text{env}}) - \alpha \mathcal{H}(g^{\sub}_i, g^{\text{ach}}_{t_i}, k)], 
    \label{unconstraint high policy update}
\end{equation}
\textcolor{black}{where $\alpha$ is a balancing coefficient to adjust the influence of reward from human guidance and distance constraint. $\mathcal{H}(g^{\sub}_{i}, g^{ach}_{t_i}, k)$ serves as a penalty function that imposes a cost when the subgoal $g^{\sub}_i$ deviates from the achievable range defined by the parameter $k$. This parameter $k$ acts as a threshold, defining the maximum allowable distance for a subgoal to be attainable. This mechanism ensures that the subgoals chosen by the high-level policy are both in harmony with human guidance and within the operational capacity of the low-level policy. }

\textcolor{black}{As the capabilities of the low-level improve, it becomes necessary to dynamically adjust the parameter $k$ to ensure that the difficulty of the subgoals remains appropriately challenging. The adjustments in $k$ lead to different optimization goals for the model, which can cause several issues: (1) The coefficient needs to strike a balance between being large enough to ensure constraints are met and small enough to avoid excessive punishment. If the penalty coefficient is set too high, it may overly restrict the solution search space, causing the algorithm to miss the optimal solution or fail to converge. Conversely, if the penalty coefficient is too low, it may not effectively enforce the constraints, leading to solutions that do not satisfy the actual problem constraints. Therefore, a static coefficient may not preserve the optimal balance between human guidance and constraints when $k$ varies. (2) Different values of $k$ lead to distinct distributions of subgoals. Sampling data pairs solely from the comprehensive offline dataset does not ensure that the reward model accurately assigns rewards to subgoals under the current $k$. (3) Assuming policy convergence based on a specific $k$, it's challenging to quickly re-optimize the high-level policy when switching to a new $k$. In response to these challenges, we have implemented three designs to ensure stability in policy updates when there are changes in $k$.}


\textbf{Automatic balancing coefficient}. We implement a dynamic balancing coefficient which will be updated in real-time to maintain a balance between the rewards and distance constraint.

To effectively incorporate the adjustment into {\M}, we need to optimize two parameters simultaneously: high-level policy $\theta_{\hi}$ and balancing coefficient $\alpha$, converting our maximization into a dual problem:

\begin{equation}
    \begin{aligned}
    \min_{\alpha \ge 0}\max_{\pi_{\hi}} \mathbb{E}_{g^{\sub}\sim \pi_{\hi}} &\sum_{i=0}^{n} [r_{\hi}(s_{t_i}, g^{\sub}_i, g^{\text{env}}) - \\ &\alpha \mathcal{H}(g^{\sub}_i, g^{\text{ach}}_{t_i}, k)].       
    \end{aligned}
\label{adjust constraint high policy update}
\end{equation}

The distance constraint function $\mathcal{H}(\cdot)$ guarantees that the subgoals within a distance $k$. This function treats subgoals that are within the distance $k$ as having zero effects. However, in the case of $\alpha$, we want it to decrease when this constraint does not work, and increase when the high-level policies make decisions that do not satisfy this constraint. Since $\mathcal{H}(\cdot)$ is always greater than 0 and the update gradient for $\alpha$ is singular, we need to eliminate the $max(\cdot)$ function from $\mathcal{H}(\cdot)$ to achieve automatic updates for $\alpha$. We update the policy $\pi_{\hi}$ firstly and then coefficient $\alpha$ following modification for the optimization:

\begin{equation}
    \begin{aligned}
        \pi_{\hi}^* = \arg\max_{\pi_{\hi}} \mathbb{E}_{g^{\sub}\sim \pi_{\hi}} &\sum_{i}^{n}[r_{\hi}(s_{t_{i}}, g^{\sub}_i, g^{\text{env}}) - \\ &\alpha \mathcal{H}(g^{\sub}_i, g^{\text{ach}}_{t_i}, k)],
    \end{aligned}
    \label{eq:policy_gradient}
\end{equation}
\begin{equation}
    \begin{aligned}
        \alpha^* &= \arg\min_{\alpha} \mathbb{E}_{g^{\sub}\sim \pi_{\hi}} \sum_{i}^{n} \left[-\alpha \left(d(g^{\sub}_i, g^{\text{ach}}_{t_i}) - k\right)\right].
    \end{aligned}
    \label{eq:alpha_gradient}
\end{equation}

\textbf{Near-policy sampling}. The probability distribution of the data varies depending on the sampling policy and the different values of $k$, even though all the data exist in an offline experience pool. It is inefficient to train the reward model using all off-policy data. Once the low-level policy has become proficient at achieving subgoals, it becomes redundant to use these subgoals for the training of the reward model, which is intended to enhance the current policy. To address this issue, we propose a new method that involves training the reward model with near-policy data, using $(s, g^{\sub}, g)$ pairs that are sampled from recent episodes. This approach allows the reward model to focus on the data that is most relevant to the current policy, enabling a more accurate evaluation of the current policy's performance. Moreover, training a reward model that can accurately evaluate the behavior of the current policy requires fewer samples and training iterations, which results in reduced computational consumption. 

Near-policy sampling can also be efficiently executed by merely preserving data from the newest episodes or their respective indices within the experience pool, and subsequently, randomly extracting from these recent datasets when gathering the comparison samples for the reward function training.

\textbf{High-level exploratory mechanism}. As depicted in Figure \ref{DDC}, the high-level policy may encounter a local optimum after proposing a subgoal in the lower-left region (left panel) and the low-level policy has adapted to achieve it, particularly upon adjusting the $k$ parameter. This policy is guided by a reward model trained on data from the replay buffer and human feedback, lacking initial pre-training. With incremental $k$ values, the high-level policy, potentially already converged, may limit its exploration to a less diverse dataset. This restricted exploration could prevent the identification and storage of superior subgoals in the replay buffer, thereby impeding the enhancement of the reward model and leading to policy training stagnation. To counteract this, it is crucial to promote high-level exploration. We have integrated exploration techniques such as RND and Maximum Entropy (MaxEnt), as detailed in \cite{mei2020global}. Specifically, we have adopted the MaxEnt approach, defining the high-level reward as $r_{\hi} = r_{\hf}(s, g^{\sub}, g^{\text{env}}) - \beta \cdot \log \pi_{\hi}(g^{\sub}|s,g^{\text{env}})$, where $\beta$ is a small constant.

\subsection{{\EED} in Low-level Policy} 
Although the high-level can provide easier goals for the low-level through RLHF, these goals also serve as guidance for the low-level to move forward. However, sparse rewards hinder the low-level ability to quickly achieve subgoals. \textcolor{black}{RLHF is well-suited for the high-level in HRL, but it is not applicable to the low-level. If the low-level policy incorporated RLHF, the noise and uncertainty in reward model could hinder the completion of the task.} Existing technologies may use hindsight relabeling, but before the low-level policy explores the subgoals, this technology cannot guarantee that the agent learns how to complete the subtasks. Due to the limited exploration ability of the low-level policy, the agent may fall into local optima and repeatedly explore meaningless areas. Introducing RND can mitigate repeated exploration by promoting novel discoveries, though its direct implementation may result in instability. RND's exploration incentives might lead to neglecting the task of subgoal completion. 

To address the issue, we propose EED, a dual-policy program, consisting of an exploration policy $\pi_{\lo}^{e}$ and a base policy $\pi_{\lo}$, both sharing the same data buffer. During interactions with the environment, we employ the exploration policy $\pi_{\lo}^{e}$ and store the experiences in the shared replay buffer. For policy updates, both the exploration and base policies undergo a hindsight relabelling process for relabeled transition data. 

Subsequently, the exploratory policy is refined by optimizing the following objective function:
\begin{align}\label{exploration policy update}
  \max_{\pi_{\lo}^{e}}& \mathbb{E}_{\tau \sim \pi_{\lo}^{e}} \left[\sum_{i=1}^{L} r_{\g} (s_{t}, g^{\sub}) + r_{\text{intri}}(s_{t})\right]. 
\end{align}
    
This formulation incorporates an additional element, $ r_{\text{intri}}(s_{t}) $, which introduces curiosity into the reward structure, thereby fostering exploration. Conversely, the base policy is updated by optimizing the following objective function:

\begin{equation}
    \begin{aligned}
        \max_{\pi_{\lo}} \mathbb{E}_{\tau \sim \pi_{\lo}^e} \left[\sum_{i=1}^{L} r_{\g} (s_{t}, g^{\sub})\right]. 
    \end{aligned}
    \label{base policy update}
\end{equation}

This approach enables the base policy to assimilate novel insights from the exploratory data without the inherent burden of the exploration process itself, thus enabling it to focus on the attainment of the defined subgoals.

\subsection{{\M} Process}
Overall, this paper introduces an innovative HRL algorithm, which proficiently identifies an approach for configuring the reward function, RLHF for high-level and EED for low-level. Furthermore, the paper addresses the challenge of inter-layer coordination in HRL by proposing a novel optimization with dynamic distance constraint. In detail, our framework {\M} works as the following pseudo-code in \Cref{alg:A}. During the interaction with the environment (from line 4 to line 17), a high-level policy $\pi_{\hi}$ is responsible for selecting a subgoal $g^{\sub}$ (line 6). Once the subgoal is determined, a low-level exploration policy, denoted as $\pi_{\lo}^e$, is utilized to execute actions until the subgoal is successfully achieved. This process of selecting subgoals and executing actions continues until the end of the episode. The data obtained from the interaction with the environment, both at the high-level and low-level, is stored in two replay buffers known as $B_{\hi}$ and $B_{\lo}$. The model update process is implemented in lines 19 to 23. The high-level policy in \Cref{alg:B} updates $\pi_{\hi}$ and $\alpha$ alternately. As for the low-level policy, it involves updating the RND model, applying hindsight, updating the low-level base policy $\pi_{\lo}$, adding the RND bonus in the batch data, and updating the exploration policy $\pi_{\lo}^e$. From lines 21 to 22, preference tuple pairs are sampled from the data of the last few episodes in $B_{\hi}$, and human labels are obtained and stored in $B_{\hf}$. These batches are then used to update $r_{\hf}$ and rewrite the reward data in the high-level buffer $B_{\hi}$. Finally, distance training data is sampled from the data of the last few episodes in $B_{\hi}$ and used to update the distance model $d$. From lines 25 to 34, it tests the low-level base policy success rate on the subgoal $g^{\sub}$ given by high-level policy and adjusts the parameters $k$.

\begin{algorithm}[tp!]
    \caption{\M}  
    \label{alg:A}  
        \begin{algorithmic}[1]
            \STATE {\textbf{Input}: initial state distribution $\rho(s)$, goal distribution $\rho(g)$, terminal threshold $\epsilon$}
            \STATE Initialize high-level policy $\pi_{\hi}$, low-level base policy $\pi_{\lo}$,  low-level exploration policy $\pi_{\lo}^{e}$,  reward model $r_{\hf}$, RND model $f_{\rnd}$, distance model $d$, balancing coefficient $\alpha$, constraint hyper-parameters $k$, high-level buffer $B_{\hi}$, preference buffer $B_{\hf}$ and low-level buffer $B_{\lo}$, a high threshold $ high\_threshold $ for the test success rate at the value $ k $, a low threshold $ low\_threshold $ for the test success rate at the value $ k $.
            \FOR{$i=0$ to $M$} 
                \STATE receive initial state $s_t$ and initial $g_{\text{env}}$ from the environment, $s_t \leftarrow \rho(s), g_{\text{env}} \leftarrow \rho(g_{\text{env}})$
                \REPEAT
                    \STATE select a subgoal $g^{\sub}$ $\leftarrow$ $\pi_{\hi}(s_t, g_{\text{env}})$
                    \STATE store high-level state $s_{\hi} \leftarrow s_t$
                    \REPEAT
                        \STATE select action $a_t \leftarrow$ $\pi_{\lo}^{e}(s_{t}, g^{\sub})$
                        \STATE execute $a_t$ and observe next state $s_{t+1}$
                        \STATE compute low-level reward $r_{\lo} = r_{\g}(s_{t+1}, g^{\sub})$
                        \STATE store $(s_t, a_t, s_{t+1}, g^{\sub}, r_l)$ to low-level buffer $B_{\lo}$ 
                        \STATE $s_{t} \leftarrow s_{t+1}$
                    \UNTIL{\textcolor{black}{$\mathbbm{1} (\| g^{\sub}-s_{t} \|_2<\epsilon)$ or end of episode}}
                    \STATE calculate the high reward $r_{\hi} \leftarrow r_{\hf}(s_{\hi}, g^{\sub}, g_{\text{env}})$
                    \STATE append $(s_{\hi}, g^{\sub}, s_{t}, g_{\text{env}}, r_{\hi})$ to $B_{\hi}$
                    \STATE $s_{\hi} \leftarrow s_{t}$
                \UNTIL{end of episode}
                \STATE \textcolor{black}{TrainHighPolicy($\pi_{\hi}$, $d$, $\alpha$, $B_{\hi}$) \Cref{alg:C}}
                \STATE TrainLowPolicy($\pi_{\lo}$, $\pi_{\lo}^e$, $f_{\rnd}$, $B_{\lo}$) \Cref{alg:B}
                \STATE near-policy query preference tuples pairs from $B_{\hi}$ and store preferences in $B_{\hf}$
                \STATE update $r_{\hf}$ on batches from $B_{\hf}$ using \Cref{reward update} and rewrite the reward in $B_{\hi}$
                \STATE near-policy sample distance batches from $B_{\lo}$ and update distance model $d$ using \Cref{distance update}
                \STATE $success\_rate \leftarrow \{\}$
                \FOR{$j=0$ to $N_1$} 
                    \STATE sample goal $g$ from environment and select a subgoal $g^{\sub}$
                    \STATE execute actions using low-level base policy$\pi_{\lo}$ on subgoal $g^{\sub}$ and store the success state in $success\_rate$          
                \ENDFOR
                \IF{average($success\_rate$) $\ge$ $high\_threshold$}
                    \STATE increase $k$ value
                \ENDIF
                \IF{average($success\_rate$) $<$ $low\_threshold$}
                    \STATE decrease $k$ value
                \ENDIF
            \ENDFOR
        \end{algorithmic}  
\end{algorithm}  

\begin{algorithm}[tp!]
\caption{TrainHighPolicy}  
\label{alg:C}  
\begin{algorithmic}[1] 
    \STATE {\textbf{Input}: high-level base policy $\pi_{\hi}$, distance model $d$, balancing coefficient $\alpha$, high-level buffer $B_{\hi}$}
    \FOR{$i=0$ to $N_2$} 
        \STATE sample a batch $B$ from buffer $B_{\hi}$
        \STATE use batch $B$ to update policy $\pi_{\hi}$ following \Cref{eq:policy_gradient}
        \STATE use batch $B$ to update dual variable $\alpha$ following \Cref{eq:alpha_gradient}
    \ENDFOR
\end{algorithmic}   
\end{algorithm}

\begin{algorithm}[tp!]
\caption{TrainLowPolicy}  
\label{alg:B}  
\begin{algorithmic}[1]
    \STATE {\textbf{Input}: low-level base policy $\pi_{\lo}$, low-level exploration policy $\pi_{\lo}^e$, RND model $f_{\rnd}$, low-level buffer $B_{\lo}$}
    \FOR{$i=0$ to $N_3$} 
        \STATE sample a batch $B$ from buffer $B_{\lo}$
        \STATE use batch $B$ to update RND model $f_{\rnd}$ using \Cref{distillation loss}
        \STATE using hindsight to rewrite the goal and reward using $r_{\g}$ in $B$
        \STATE use batch $B$ to update policy $\pi_{\lo}$ using \Cref{base policy update}
        \STATE add RND bonus $f_{\rnd}$ to reward in $B$ using \Cref{intrinsic reward}
        \STATE use batch $B$ to update exploration policy $\pi_{\lo}^e$ using \Cref{exploration policy update}
    \ENDFOR
\end{algorithmic} 
\end{algorithm}

\begin{figure*}[!htbp]
    \centering
    \begin{minipage}[t]{0.15\textwidth}
        \centering
        \includegraphics[width=0.9\linewidth]{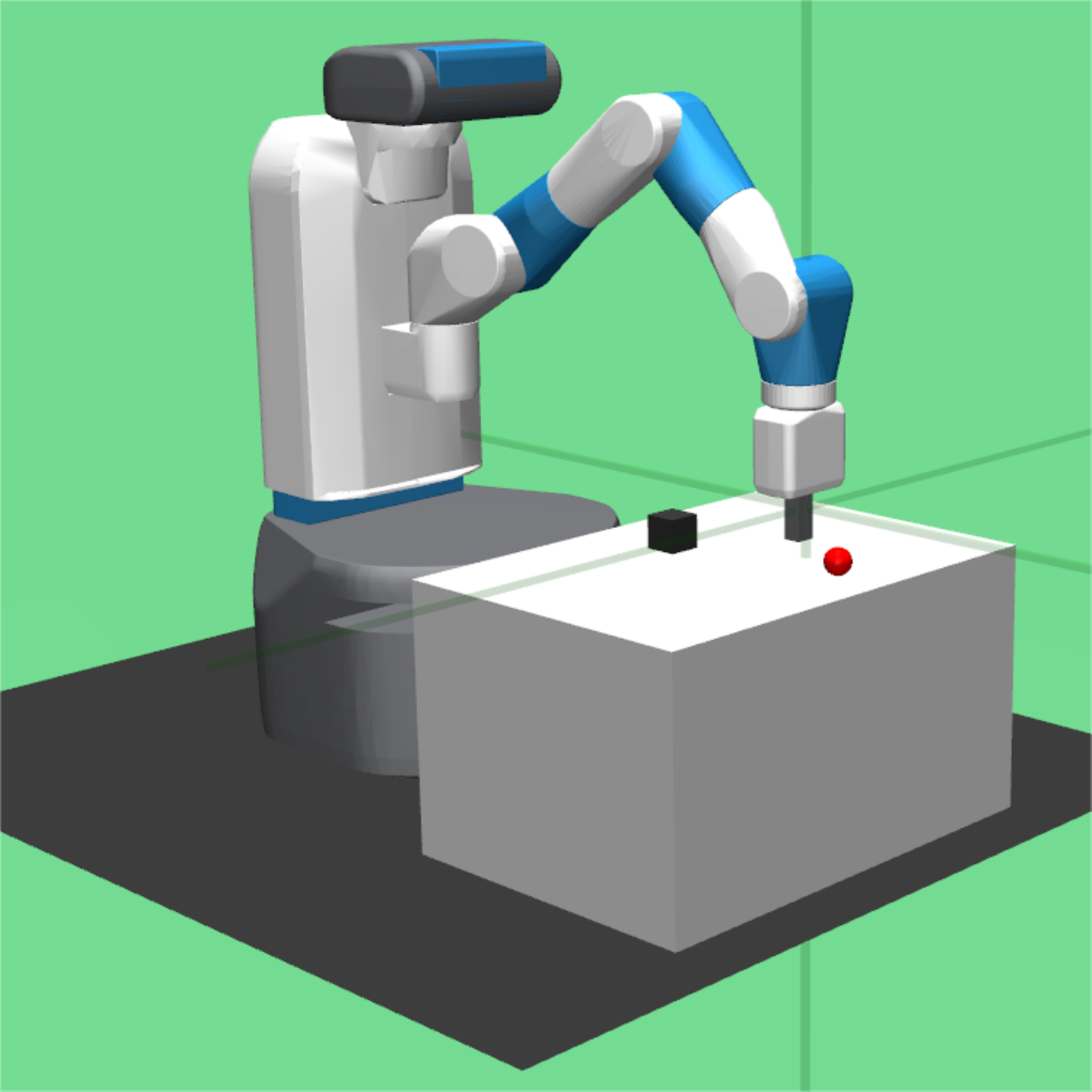}
        \caption*{\footnotesize FetchPush}
    \end{minipage}
    \begin{minipage}[t]{0.15\textwidth}
        \centering
        \includegraphics[width=0.9\linewidth]{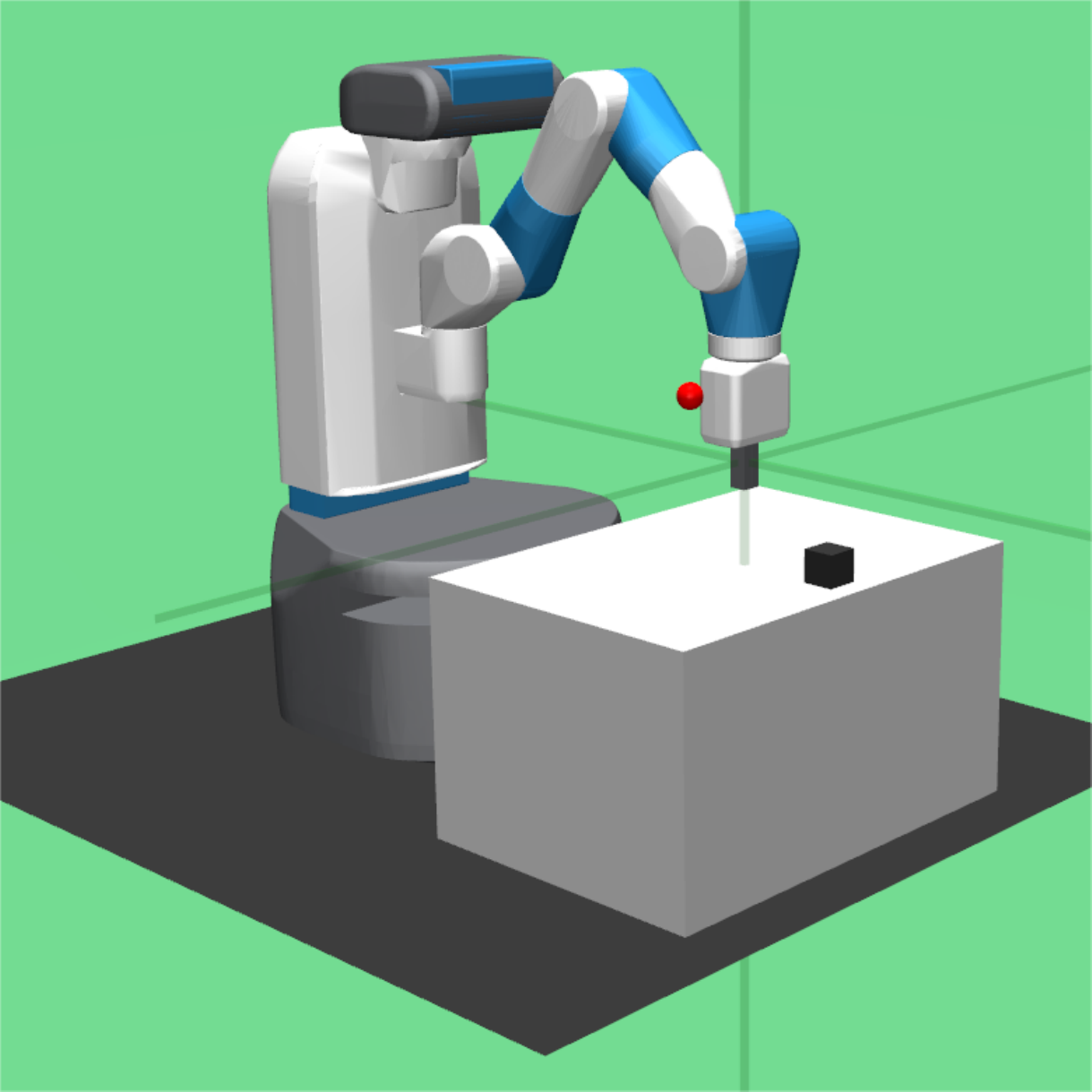}
        \caption*{\footnotesize FetchPickAndPlace}
    \end{minipage}
    \begin{minipage}[t]{0.15\textwidth}
        \centering
        \includegraphics[width=0.9\linewidth]{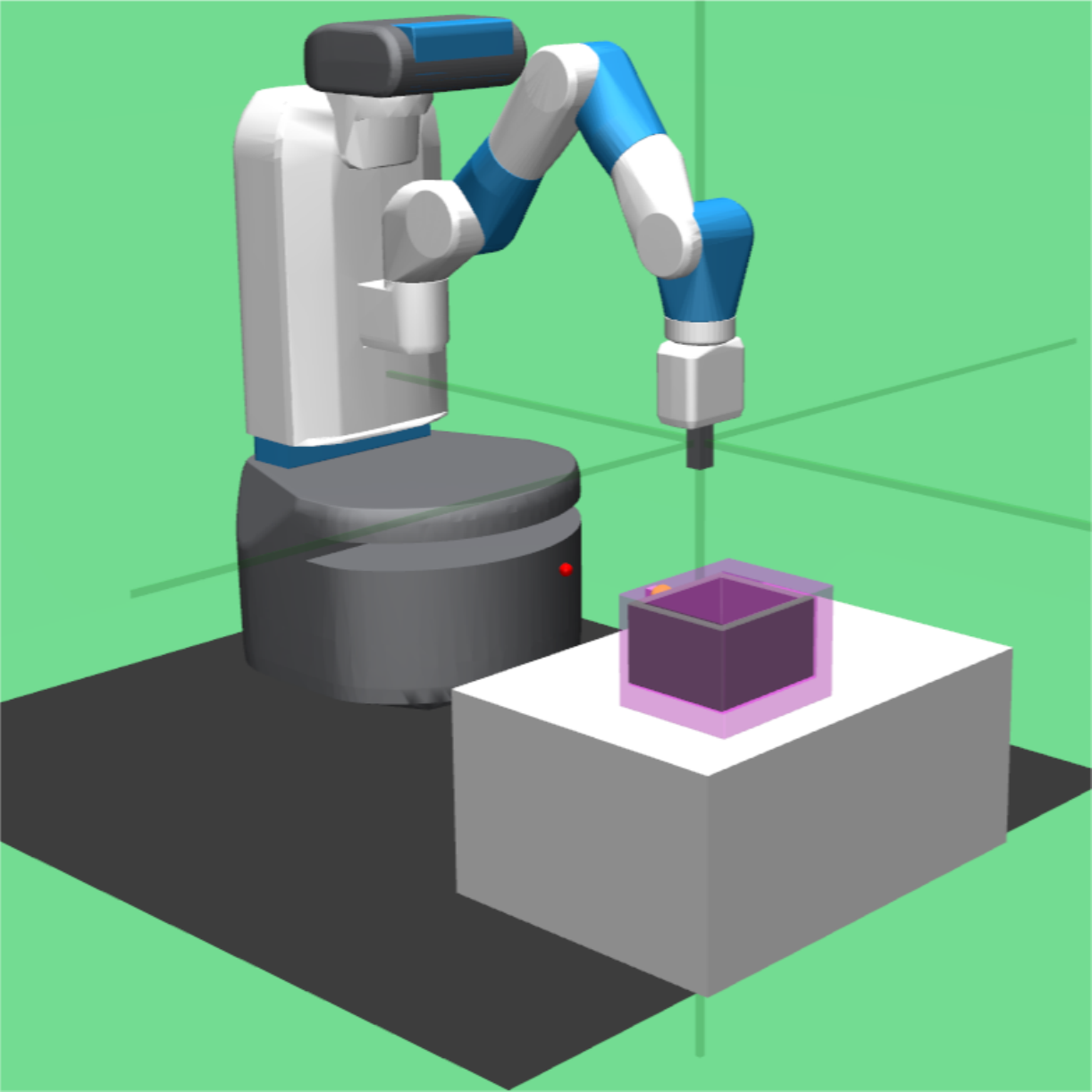}
        \caption*{\footnotesize FetchDraw}
    \end{minipage}
    \begin{minipage}[t]{0.15\textwidth}
        \centering
        \includegraphics[width=0.9\linewidth]{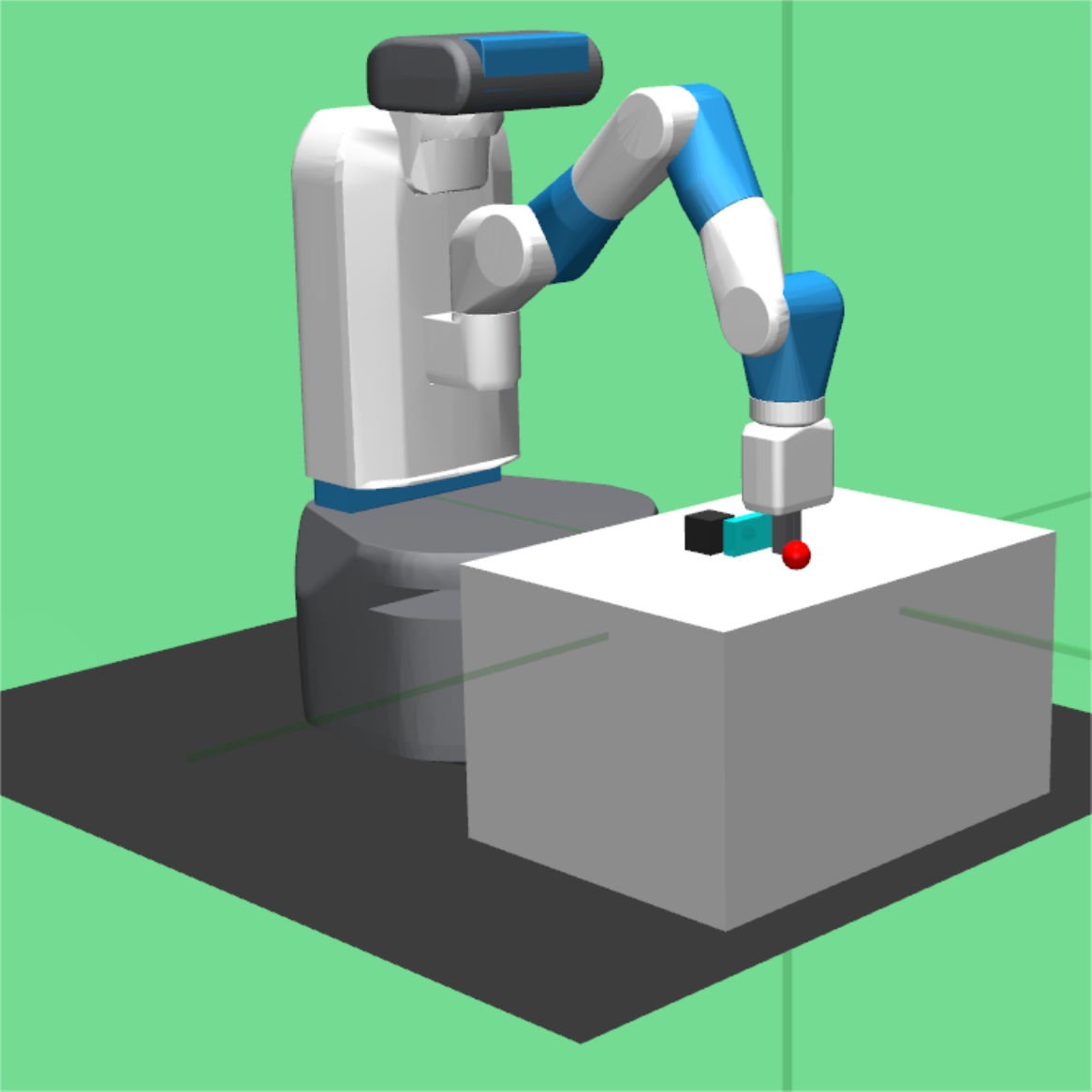}
        \caption*{\footnotesize FetchObsPush}
    \end{minipage}
    \begin{minipage}[t]{0.15\textwidth}
        \centering
        \includegraphics[width=0.9\linewidth]{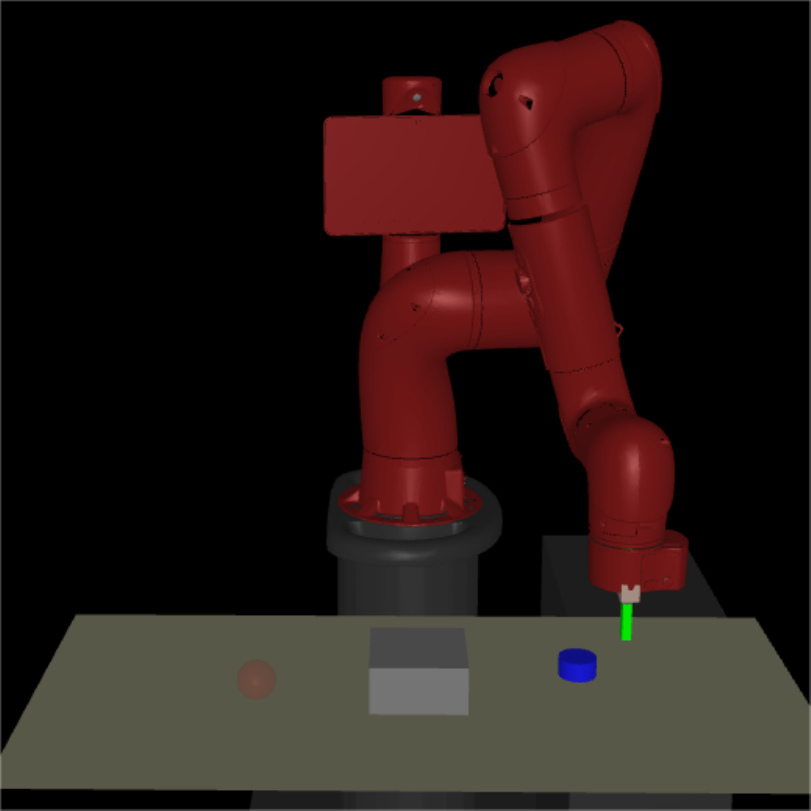}
        \caption*{\footnotesize Pusher}
    \end{minipage}
    \begin{minipage}[t]{0.15\textwidth}
        \centering
        \includegraphics[width=0.9\linewidth]{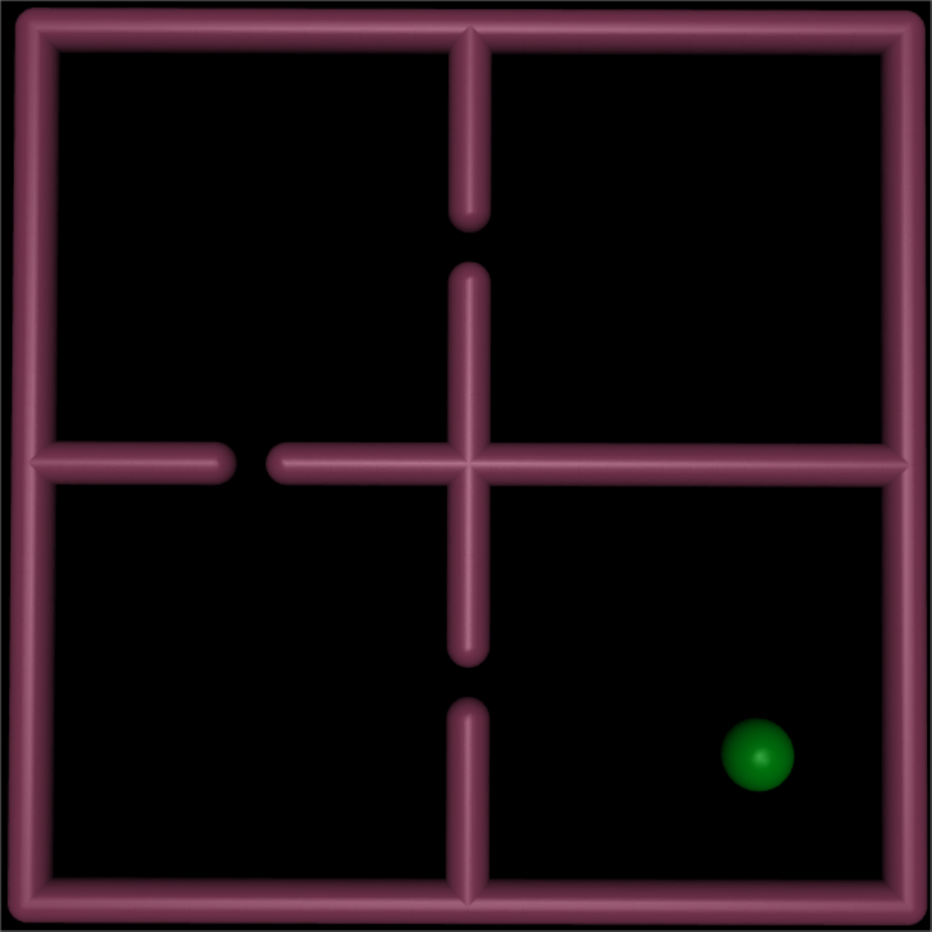}
        \caption*{\footnotesize Four rooms}
    \end{minipage}
    
    \caption{\textcolor{black}{Experimental Domains for Evaluating {\M} compared with baselines. The Pusher and Four Rooms serve as representative domains within the class of Partially Observable Markov Decision Processes (POMDPs), characterized by uncertainties in the state observations which require the algorithm to make decisions based on incomplete information. In contrast, the rest domains, such as FetchObsPush, provide a fully observable state space. For further details, please refer to \Cref{appendix:B}.}
    }
    
    \label{domain}
\end{figure*}

\section{EXPERIMENTS}
In this section, we perform a range of experiments in various commonly used domains \cite{plappert2018multi, luo2023relay, ghosh2019learning}, as depicted in \Cref{domain}. Through the experiments, we aim to answer the following research questions (RQs):
\begin{description}
    \item[RQ1] How does the performance of {\M} in comparison to baseline models in various domains?
    \item[RQ2] What is the impact of DDC in {\M}?
    \item[RQ3] What is the impact of human feedback in {\M}?
    \item[RQ4] What insights can be gained about the individual contributions of key components in {\M}?
\end{description}

\subsection{Setup}
\textbf{Benchmarks:} We select FetchPush, FetchPickAndPlace, FetchDraw, FetchObsPush, Pusher, and Four rooms as our simulation benchmarks, widely used in research \cite{andrychowicz2017hindsight, bougie2022hierarchical, ghosh2019learning}. As illustrated in \Cref{domain}, the first five domains involve long-horizon manipulation tasks, while Four rooms focuses on 2D navigation. In our experiments utilizing the RLHF, we employed two distinct approaches to acquire human labels: manual labeling by humans and synthetic labeling through scripts. The specifics of these label acquisition processes are delineated in the \Cref{RLHF details}. Across the experimental trials, we primarily utilized synthetic labels, with the exceptions explicitly noted.

\textbf{Hardware:} The computer's hardware is equipped with an AMD Ryzen 7 5700X processor, 32GB of RAM, and an NVIDIA GeForce RTX 3070 Ti graphics card.

\textbf{Baselines:} We have incorporated various learning methods in our baseline implementation, including techniques from HRL, RLHF, RND, dynamic distance learning, and hindsight relabelling.

\begin{itemize}
\setlength{\itemsep}{0pt}
\setlength{\parsep}{0pt}
\setlength{\parskip}{0pt}
\item \textbf{Reinforcement Learning from Human Feedback (RLHF)} firstly establishes a reward model based on human feedback and then using it to optimize policy in reinforcement learning. \textbf{PEBBLE} is a RLHF framework that that combines unsupervised pre-training and preference-based learning to significantly improve the sample and feedback efficiency of human-in-the-loop reinforcement learning \cite{lee2021pebble}. The key concept of PEBBLE is to learn a reward model on human preference. Unlike the learning strategy of the reward model in the {\M}, it has an unsupervised pretraining step via intrinsic reward. Here, we apply the state entropy $H(s) = - \mathbb{E} \log p(s)$. By converting the policy to a simpler version, the pretraining reward function is set as $\text{log}(||s_{i}-s_{i}^k||)$ in the batch. This implies that maximizing the distance between a state and its nearest neighbor increases the overall state entropy. 
\item \textbf{Hindsight Relabeling} is a data augmentation method for multi-task reinforcement learning that facilitates data sharing across tasks by relabeling experiences to improve sample efficiency. Our baseline, \textbf{HER} \cite{andrychowicz2017hindsight} is a classical hindsight relabeling strategy. HER enables efficient learning from limited and non-diverse rewards by re-framing failed attempts as successes towards different goals. In our implementation, we utilize a policy to incorporate HER. When sampling batch transitions from the replay buffer, we employ hindsight technology to modify certain parts of the transitions. It is important to mention that the high-level policy differs from the low-level policy of {\M} in certain aspects. Unlike the low-level policy, the high-level policy solely receives goals from the environment and does not incorporate an RND bonus. 
\item \textbf{Hierarchical Reinforcement Learning with Human Feedback (HRL-HF)} integrates human feedback into the generation of high-level subgoals. We follow the architecture of \textbf{HhP} \cite{bougie2022hierarchical}. HhP introduces an HRL method that integrates human preferences and curiosity-driven exploration. By using human preferences to guide decision-making at high-levels and curiosity to promote environmental exploration at low-levels. HhP is considered to be inefficient in combining HRL and RLHF, and it also introduces bias at the low-levels. When we trained HhP, we found that this algorithm was difficult to converge, and to ensure the effectiveness of training, we introduced the hindsight relabelling technique in this algorithm as well. 
\item \textbf{Distance Learning (DL)} learns distance models. The model is utilized to train goal achievement by setting the reward as the negative distance between states and goals. This baseline follows \textbf{DDL} \cite{hartikainen2019dynamical}. DDL calculates dynamical distances, defining them as the expected number of steps between states in a system. This method leverages unsupervised interactions for learning distances. In the distance evaluation step, the aim is to estimate the dynamic distance between pairs of states visited by a given policy. A distance function, denoted as $d(s_i, s_j)$, is utilized for this purpose. To calculate the distance, multiple trajectories, denoted as $\tau $, are generated by rolling out the policy. Each trajectory has a length of T. The empirical distance between states $s_i$ and $s_j$ in the trajectory $\tau$ is then calculated, where $0 \leq i \leq j \leq T$. The empirical distance is simply given by the difference between j and i. To learn the distance function, supervised regression can be employed by minimizing the objective: 

$\mathcal{L}_d = \frac{1}{2} \mathbb{E}_{\tau \sim p_\tau} \left[ \sum_{i \in [0,T]} \sum_{j \in [i,T]} \left( d_{\psi}(s_i, s_j) - (j - i) \right)^2 \right]$.
\end{itemize}

\begin{figure*}[!htbp]
    \centering
    \includegraphics[width=0.8\linewidth]{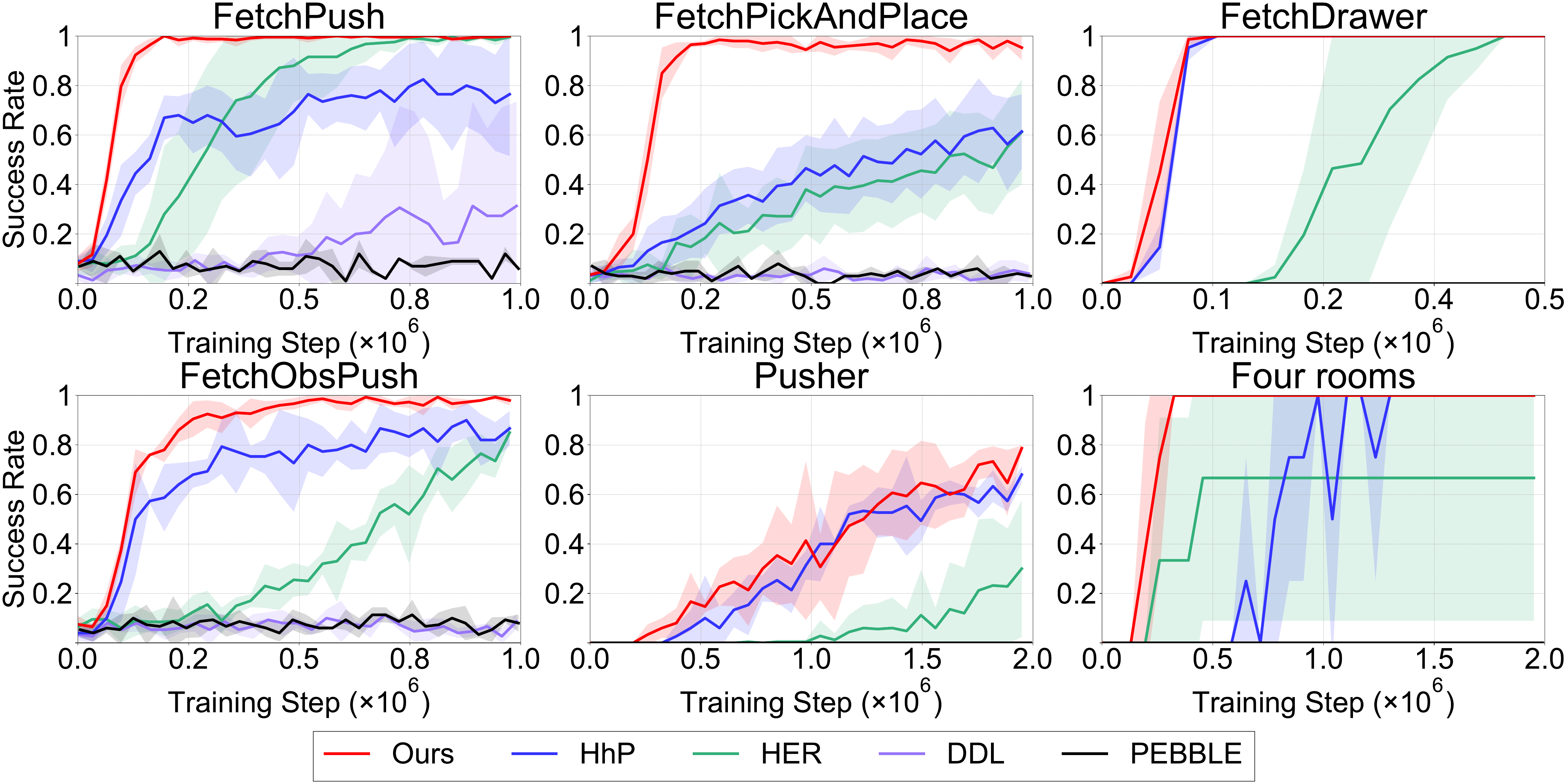}
    \caption{Graphical representation of the success rates for {\M} in comparison to other baseline methods across different benchmarks on five random seeds. The shaded areas surrounding each curve represent the standard deviation. \textcolor{black}{Within the Four Rooms domain, the performance curve exhibits non-smooth behavior due to the fixed positions of the starting point and the goal. Consequently, the success rate can abruptly transition from 0\% to 100\%, leading to the curve with large variance.} Any curves that are not visible in the graph, indicate a zero success rate throughout the trials. These results are aggregated from an average of five individual runs.}
    \label{exp1}
\end{figure*}

\subsection{Performance Evaluation (RQ1)}
In our experiments across six domains, we find that {\M} excels in learning speed and subgoal attainment as evidenced in \Cref{exp1}. This assessment uses five random seeds and is evaluated over 50 tests. As can be seen from DDL and PEBBLE curves. DDL and PEBBLE rarely learn effectively in complex GCRL environments. DDL's poor performance may be attributed to the absence of guiding signals and the instability of the reward function. Pairwise comparison guidance makes it difficult for PEBBLE to complete subgoal. HER, a classic GCRL algorithm, serves as a benchmark for evaluating other algorithms. Yet, HhP, despite integrating human guidance and curiosity-driven exploration, underperforms compared to HER in FetchPush. Its benefits are also unclear in Pusher and Four Rooms. This indicates HhP's inadequate use of human guidance and curiosity in exploration. In contrast, {\M}, by incorporating {\DDC} and {\EED}, effectively leverages these elements for utilization of human feedback and exploration, achieving faster and more consistent training than DDL, PEBBLE, and HER. Our analysis highlights the superior performance of {\M}. 

\begin{table}[tp!]
\centering
\caption{\textcolor{black}{Model parameter count and time consumption for the first 1000 episodes on FetchPush domain. Here, 'M' denotes 'million' for the quantity of model parameters, and 'min' denotes 'minutes' for the duration of time spent.}}
\label{tab:time-analysis}
\begin{tabular}{c|lll}
\hline
{Algorithm} & \textbf{MENTOR}  & \textbf{HER}\\
\hline
low-level actor\&critic & 0.560M & 0.280M \\
high-level actor\&critic & 0.189M & 0.189M \\
reward & 0.140M - \\
rnd & 0.278M & - \\
distance & 0.068M & -\\
total count & 1.235M & 0.469M\\
time consumption & 18.63$\pm$ 0.05min & 10.76$\pm$ 0.08min \\
\hline
\end{tabular}
\end{table}

\textcolor{black}{However, MENTOR demonstrates superior performance and is equipped with a greater number of components. To investigate the resource consumption of our algorithm, we conducted additional experiments, meticulously recording the model parameter count and computation time for 1000 iterations of MENOTR and HER in the FetchPush environment. As detailed in \Cref{tab:time-analysis}, our results reveal that MENTOR has approximately 263\% more parameters and requires roughly 176\% more running time compared to HER. Although our model has a larger number of components, which significantly increases the parameter count, considering that modern GPUs typically have 8GB or more of memory, these models do not demand excessively high computational resources. Moreover, by examining the convergence of HER in \Cref{exp1} and the computational time consumed by both MENTOR and HER algorithms over 1000 episodes, we found that our algorithm consumes 76\% more time. However, the convergence speed of our algorithm is more than 3 times that of HER. Therefore, our algorithm remains highly efficient in terms of resource consumption.}

\subsection{Impact of DDC (RQ2)}
Our study will examine DDC's effects on learning agents, assessing its impact on agent behaviors and the efficacy of three designs in achieving stable algorithmic convergence. We also explore DDC's synergism with human feedback.

\begin{figure}[!htbp]
    \centering
    \begin{minipage}[t]{0.47\textwidth}
        \centering
        \includegraphics[width=\linewidth]{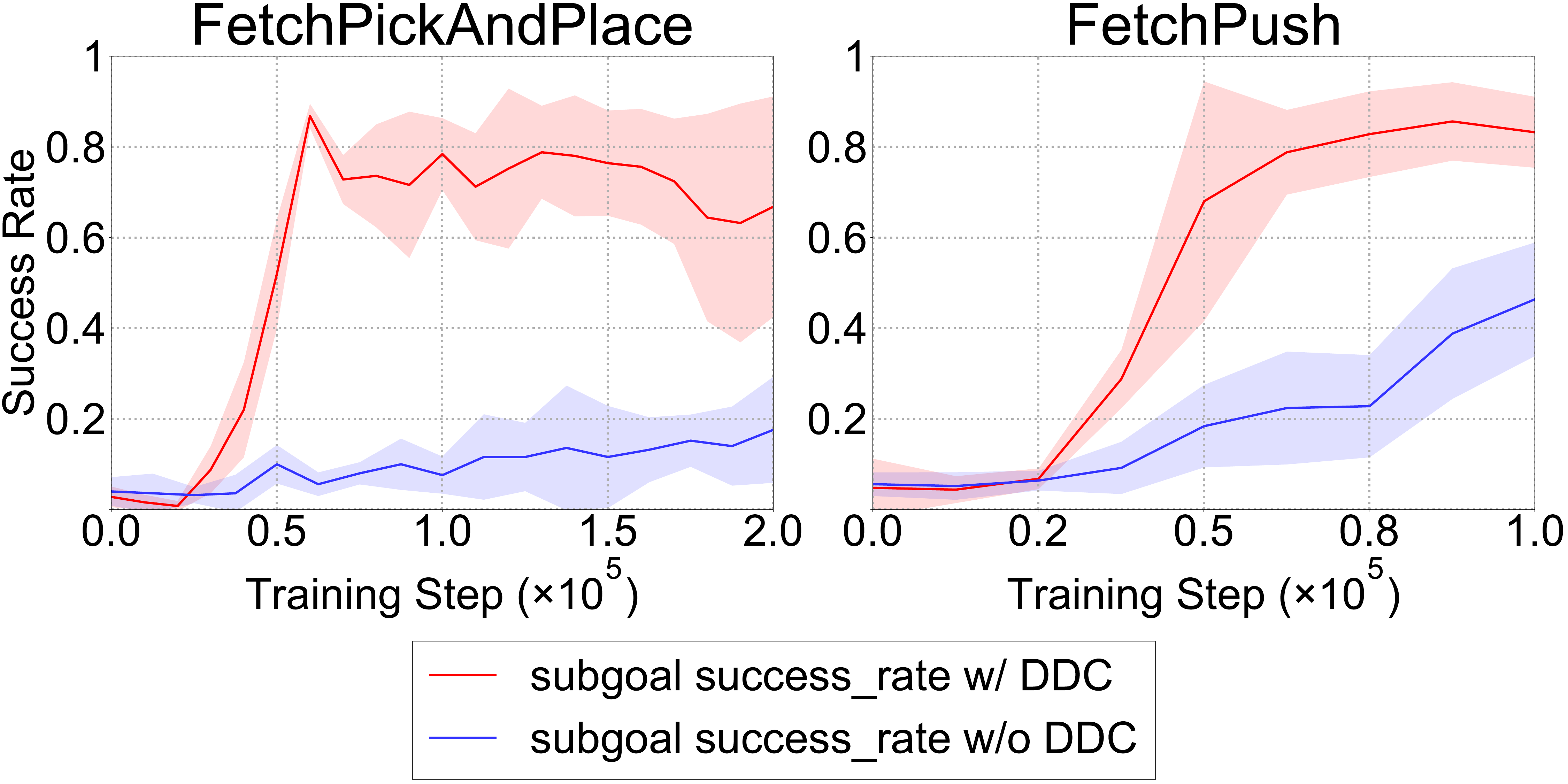}
    \end{minipage}
    \begin{minipage}[t]{0.47\textwidth}
        \centering
        \includegraphics[width=\linewidth]{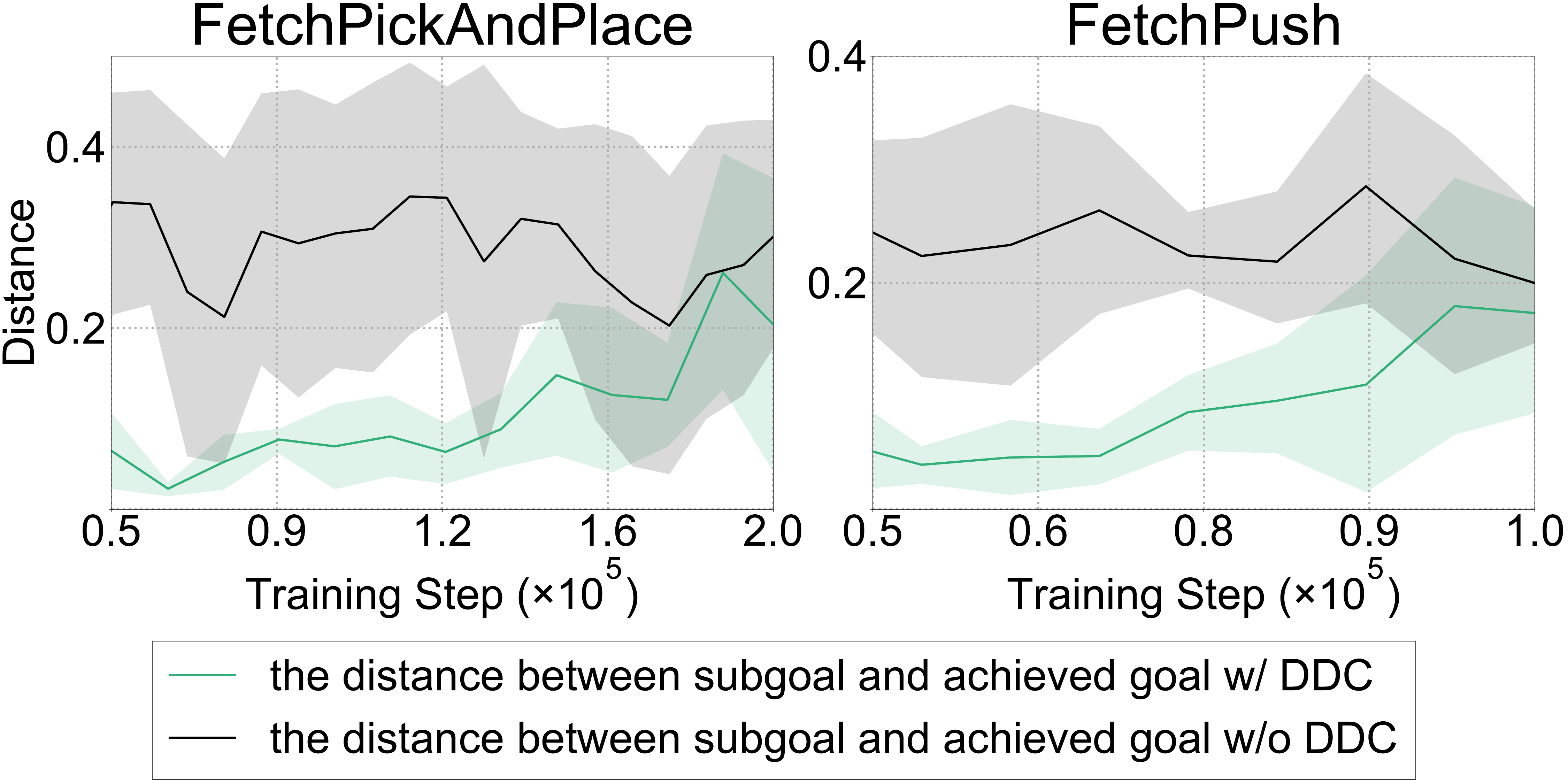}
    \end{minipage}
    \caption{Impacts of Distance Constraints on success rate in FetchPickAndPlace and FetchPush domains. Since the high-level policy requires data to be collected before updating, a segment is missing from the distance curve.}
    \label{exp2}
\end{figure}

\textbf{Examining the correlation between task completion and DDC}. As illustrated in \Cref{exp2}, when comparing the black and green curves on the lower side, It is evident that DDC can regulate the difficulty of subgoals provided by limiting the distance. Without DDC, the high-level policy proposes subgoals at random difficulty. By examining this phenomenon in conjunction with the success rate of the subgoals, we can draw the following conclusions: (1) during the initial stages of the training process, the low-level policy can rapidly acquire the ability to achieve easy subgoals. (2) Once the low-level policy has successfully mastered a subgoal of a certain difficulty level, it can efficiently progress to learning subgoals of slightly higher difficulty. When there is no DDC, subgoals of randomized difficulty lead to slower learning of the low-level policy. It is concluded that the DDC can efficiently coordinate high-level and low-level.

\begin{figure}[tp!]
    \centering
    \includegraphics[width=\linewidth]{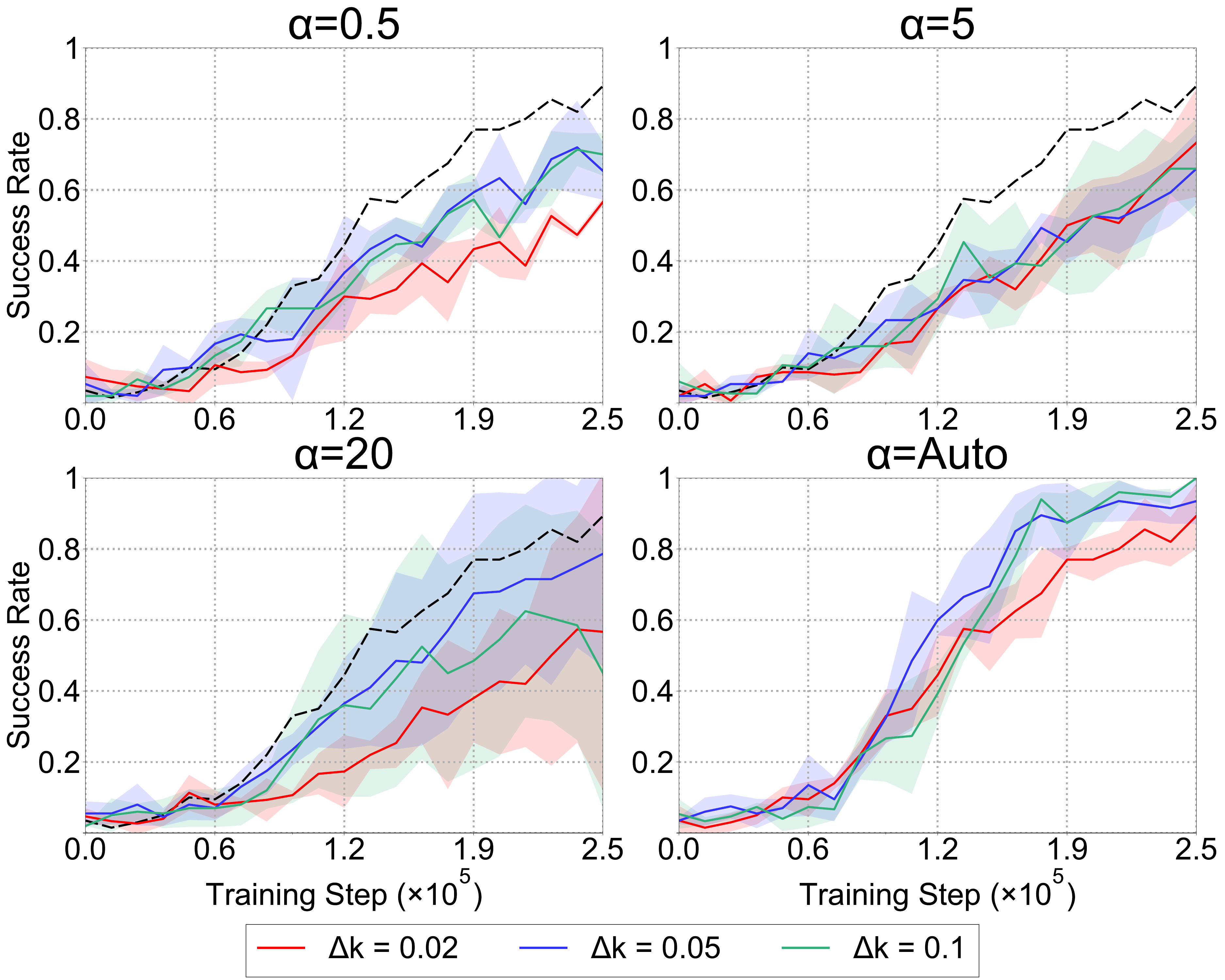}
    \caption{Effects of the balancing coefficient on the environment goal success rate in FetchPickAndPlace domains are examined on five random seeds. In the first three graphs, the dashed lines represent the average success rate with auto-set $\alpha$ in the worst case, where $\Delta k$ is 0.02. The adjustment value of $k$ is represented as $\Delta k$. We modify the parameter $k$ to increase on successful completion of the subgoal and decrease on failure.}
    \label{dynamicalpha}
\end{figure}

\textbf{Analyzing the impact of automatic balancing coefficient}. DDC has a mechanism to automatically adjust the balancing coefficient. The mechanism plays a crucial role in balancing the impact of the reward model and distance constraint. As shown in \Cref{dynamicalpha}, fixed values of 0.5, 5, and 20 for $\alpha$ are ineffective in learning compared to the automatic adjustment setting. Small $\alpha$ values (0.5 and 5) slow DDC convergence. Large $\alpha$ (20) enhances learning speed but makes it unstable and sensitive to $\Delta k$. 

\begin{figure}[tp!]
    \centering
    \includegraphics[width=0.8\linewidth]{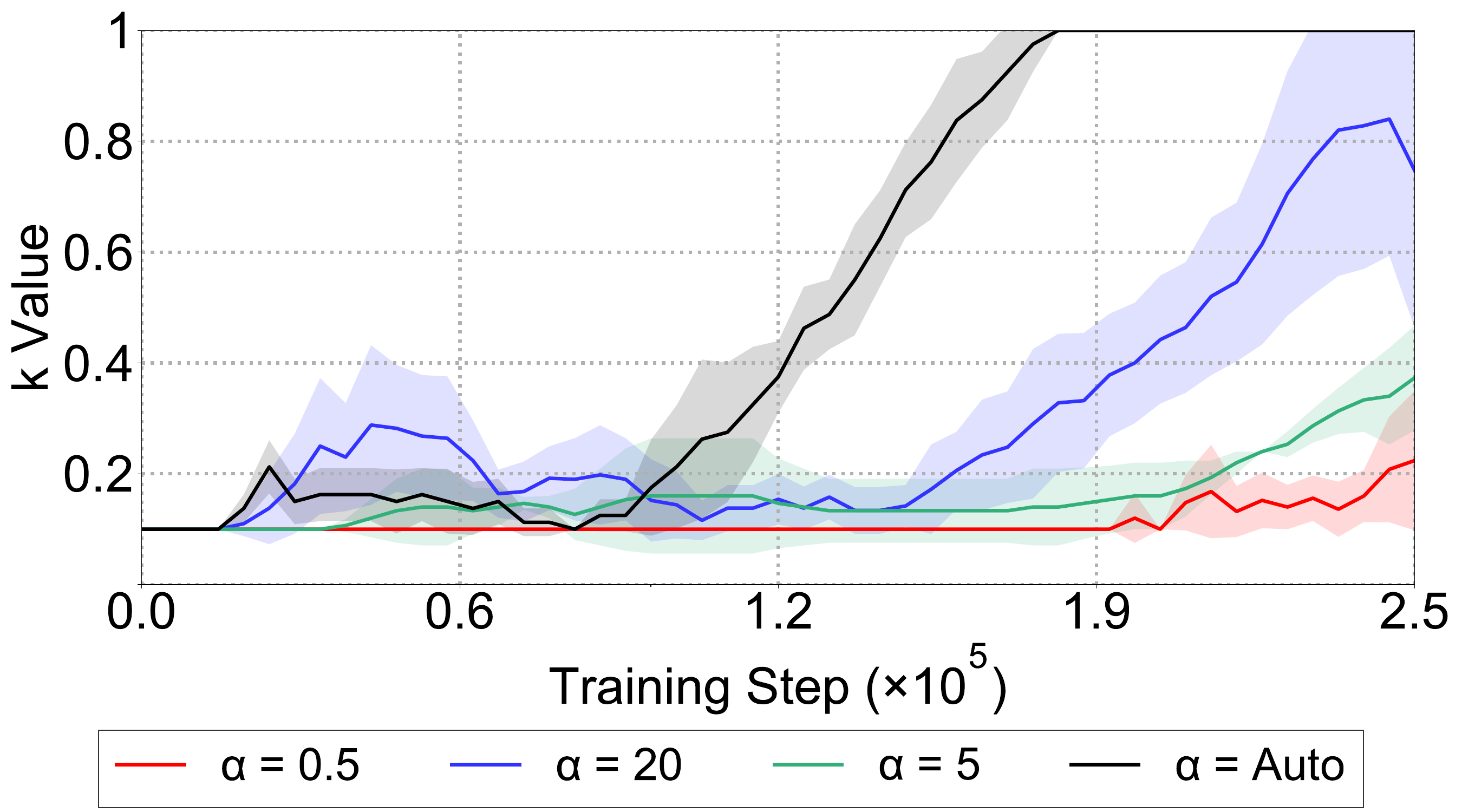}
    \caption{\textcolor{black}{The variation curve of $k$ value throughout the training process under different $\alpha$ value and $\Delta k$ is 0.05.}}
    \label{dynamick}
\end{figure}

\textcolor{black}{To gain more insight into the details of difficulty adjustment under different $\alpha$, we recorded the change curve of $k$ ($\Delta k$ of 0.05) under different $\alpha$ settings in \Cref{dynamick}. Analyzing the $k$ value change curves for different $\alpha$ values reveals how subgoal difficulty adapts throughout training. These curves track the incremental changes in the $k$ value ($\Delta k$ = 0.05) and highlight the influence of the $\alpha$ parameter on the stability and progression of the learning trajectory. Lower $\alpha$ values lead to a more gradual increase in the $k$ value, promoting a stable but slower learning process. In contrast, higher $\alpha$ values cause more pronounced fluctuations, which may accelerate learning but also introduce greater risk of instability. The Auto-$\alpha$ setting demonstrates both rapid and stable $k$ value growth, indicating an adaptive strategy that balances speed and stability in the learning process. These $k$ value curves also serve as indicators of the low-level policy's progress: consistent increases suggest ongoing improvement, while plateaus or declines may signal the need to reassess the strategy. These analyses highlight the crucial role of the $\alpha$ parameter in shaping the learning process.}

\begin{figure}[tp!]
    \centering
    \includegraphics[width=\linewidth]{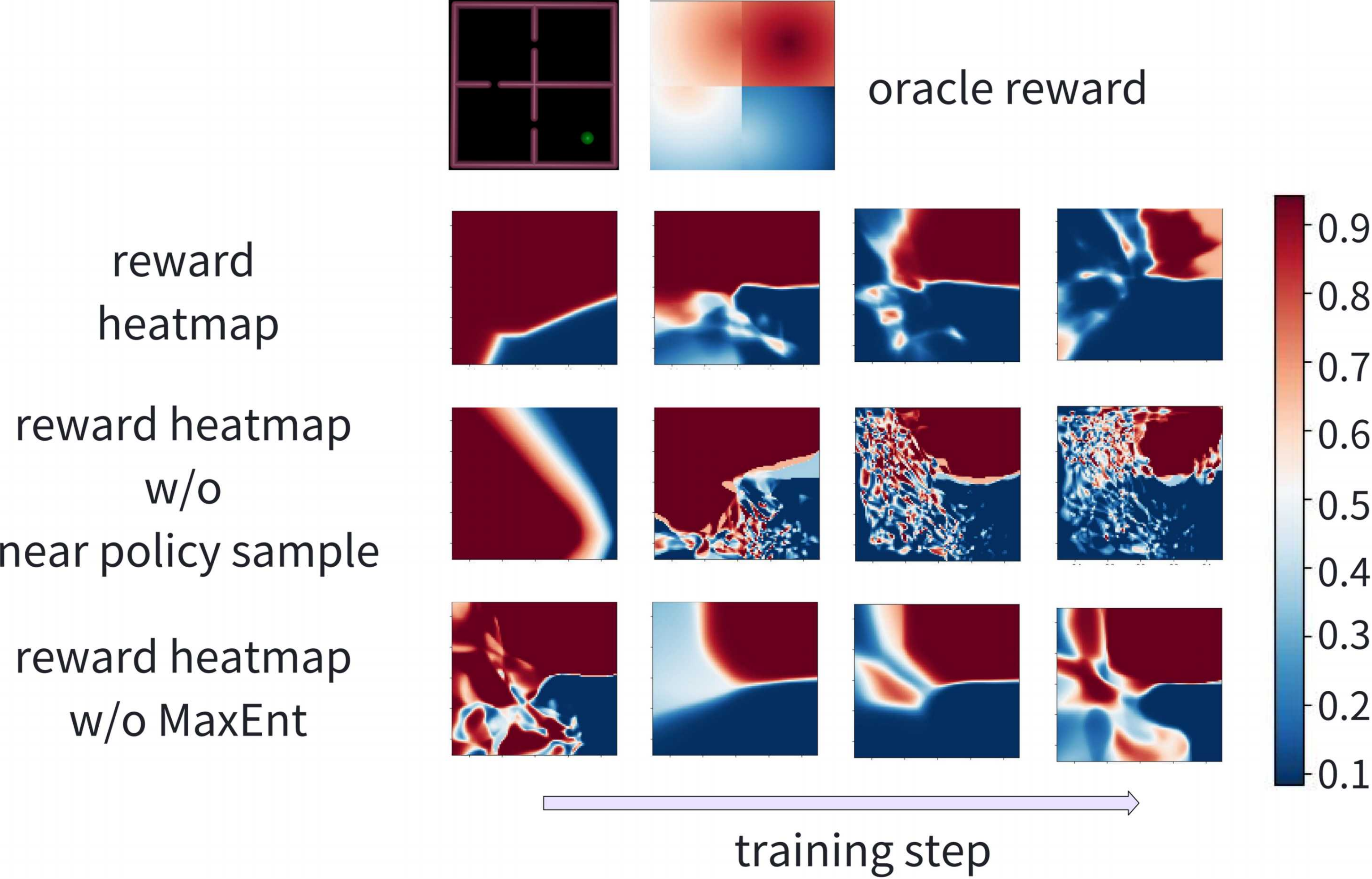}
    \caption{\textcolor{black}{Near-policy sampling and exploratory mechanism effects on reward model in the Four rooms domain. On the top, the oracle reward function in the Four rooms domain indicates where the agent likely earns higher rewards, guiding it from the lower right to the upper right corner. We use the oracle reward to generate synthetic labels to train the reward model following \Cref{reward update}. The heatmaps below the oracle reward reflect how the reward model adjusts in response to the agent's exploration and policy updates within the environment. The heatmaps of reward model for the top and bottom exhibit changes across 2,000 episodes, whereas the heatmap for the center demonstrates alterations throughout 10,000 episodes.}}
    \label{appendix exp2}
\end{figure}

\textbf{Analyzing the near-policy sampling for reward model training and the exploratory mechanism at the high-level}. During the training process of {\M}, we observe some phenomena that the high-level policy cannot be well guided by the rewarding model. These observations can be ascribed to two factors: (1) the reward model's inadequate guidance of the current model, and (2) the increase in the $k$ value leading to the agents in local optima, thereby diminishing further exploration of potential subgoals. In \Cref{appendix exp2}, the reward heatmap, shaped by human feedback, approximates the oracle function. We observe that when near-policy samples are absent, the reward model tends to emphasize the entirety of the state space. This, in turn, makes it more challenging and computing resource-consuming for the reward model to effectively learn. 

Additionally, we have observed that there are instances where the high-level policy becomes stuck in local optima as we incrementally increase the value of $k$ for a long time. This phenomenon is attributed to two reasons: (1) it lacks pretraining at the high-level, and (2) the high-level lacks the exploratory ability, leading to not trying new subgoals. Pretraining plays a crucial role in RLHF, yet it can be inefficient at the high-level due to limited samples (often only one per episode). In the operation of the high-levels, we find it necessary to increase the exploratory ability. As shown in \Cref{appendix exp2}, reward learning becomes challenging without MaxEnt (maximum entropy). We compare RND and MaxEnt in terms of exploratory at high-levels. After conducting experiments displayed in \Cref{appendix explore_ablation}, we discovered that MaxEnt outperforms RND. While RND promotes the exploration of novel states, MaxEnt focuses on improving action entropy. The combination of the reward model and DDC can lead to previously disregarded decisions being recognized as good decisions. RND can learn and incorporate these decisions previously, treating them as non-novel at late times. However, MaxEnt does not encounter this issue and as a result, it achieves significantly better performance compared to RND.

\begin{figure}[tp!]
    \centering
    \begin{minipage}[t]{0.2\textwidth}
        \centerline{\includegraphics[width=\linewidth]{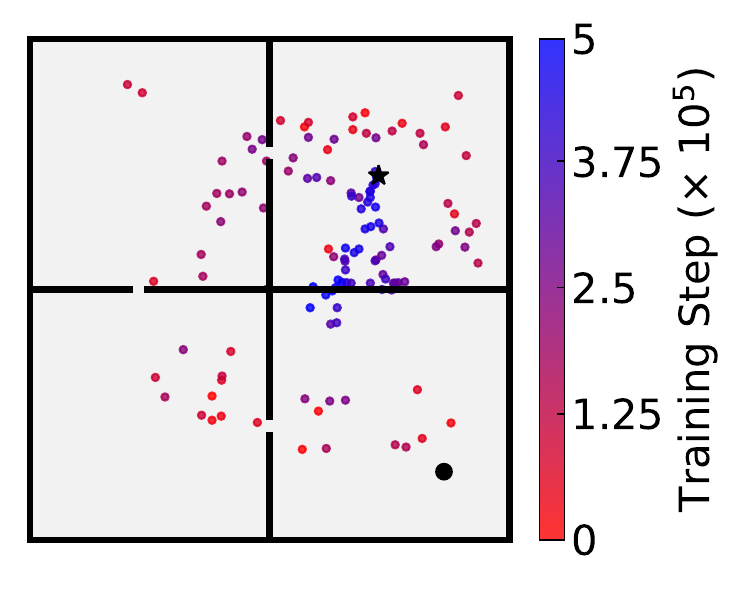}}
        \centerline{\footnotesize w/ DDC}
    \end{minipage}
    \hfill
    \begin{minipage}[t]{0.2\textwidth}
        \centerline{\includegraphics[width=\linewidth]{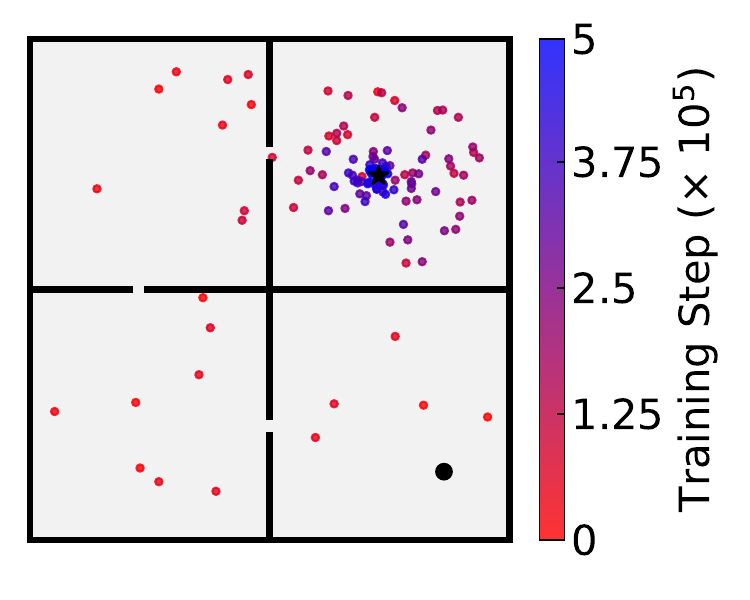}}
        \centerline{\footnotesize w/o DDC}
    \end{minipage}
    \caption{In Four rooms domain, we compare subgoal distributions with and without DDC during training. Subgoals are shown as colored circles, with a red-to-blue gradient for training time. The starting point is marked by a black circle in the lower right, while the ending point is a pentagram in the upper left.}
    \label{backgroundddc}
\end{figure}

\begin{figure}[tp!]
    \centering
    \includegraphics[width=0.8\linewidth]{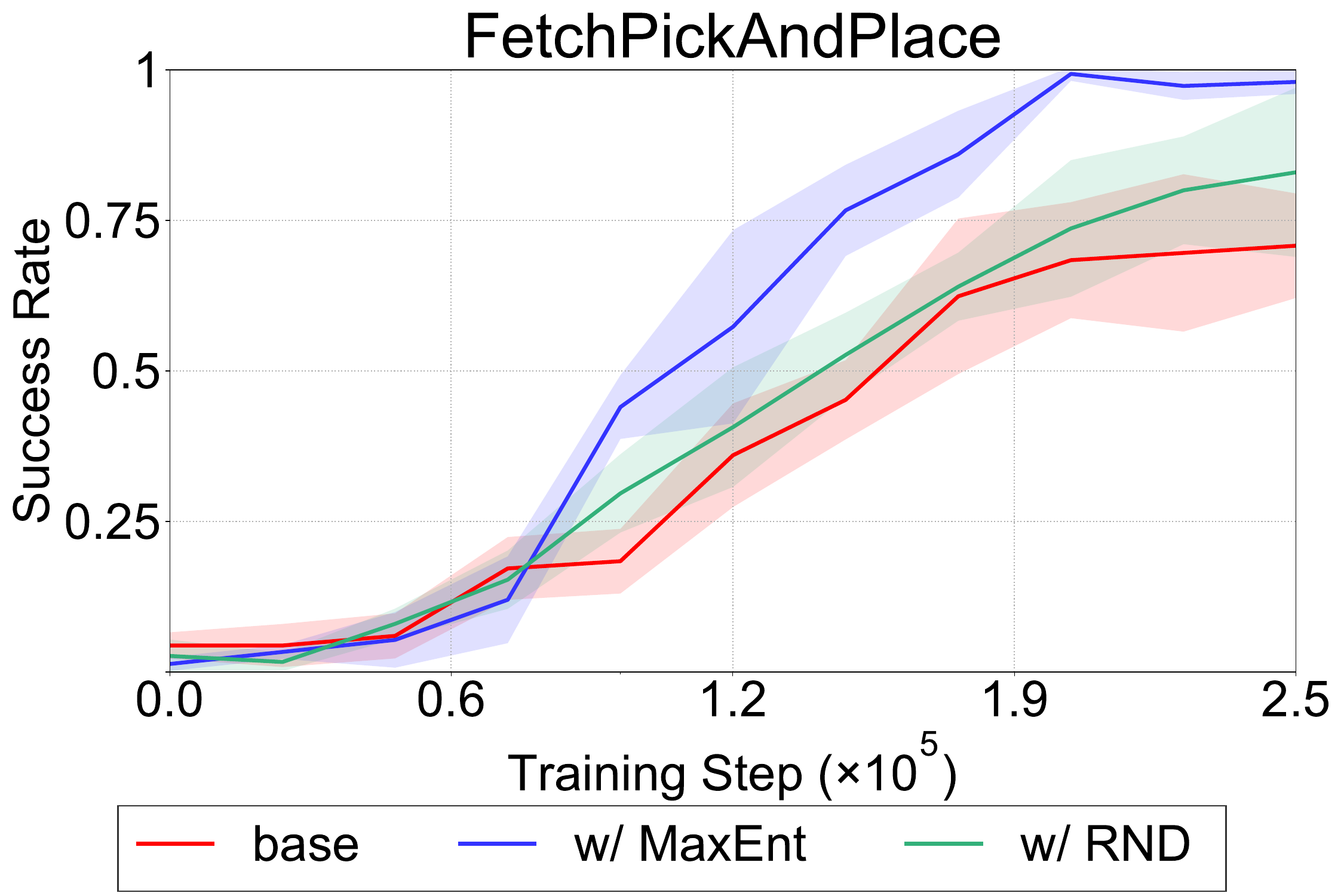}
    \caption{Exploratory mechanism effects analysis in FetchPickAndPlace domain.}
    \label{appendix explore_ablation}
\end{figure}

\textbf{Analyzing the overlay effects of DDC and human feedback}. Having investigated DDC's impact, we now turn to assess the combined influence of human guidance and DDC on subgoal formulation. \Cref{backgroundddc} depicts the training phase subgoal distribution, both with and without DDC. To the right of the figure, DDC's inactivity results in a fragmented learning interaction between levels. While the high-level, aided by human guidance, swiftly navigates to the goal, it neglects the low-level's execution capabilities, leading to transient and inefficient guidance, evident in the predominance of red subgoals in most regions except the upper-right corner. In contrast, on the figure's left, with DDC active, the high-level delineates a subgoal sequence starting from the lower right, proceeding to the lower left, then the upper left, and culminating at the upper right corner. This strategy, different from the scenario without DDC, significantly bolsters the efficiency of human guidance, contributing to {\M} overall effectiveness. 

\begin{figure}[tp!]
    \centering
    \includegraphics[width=\linewidth]{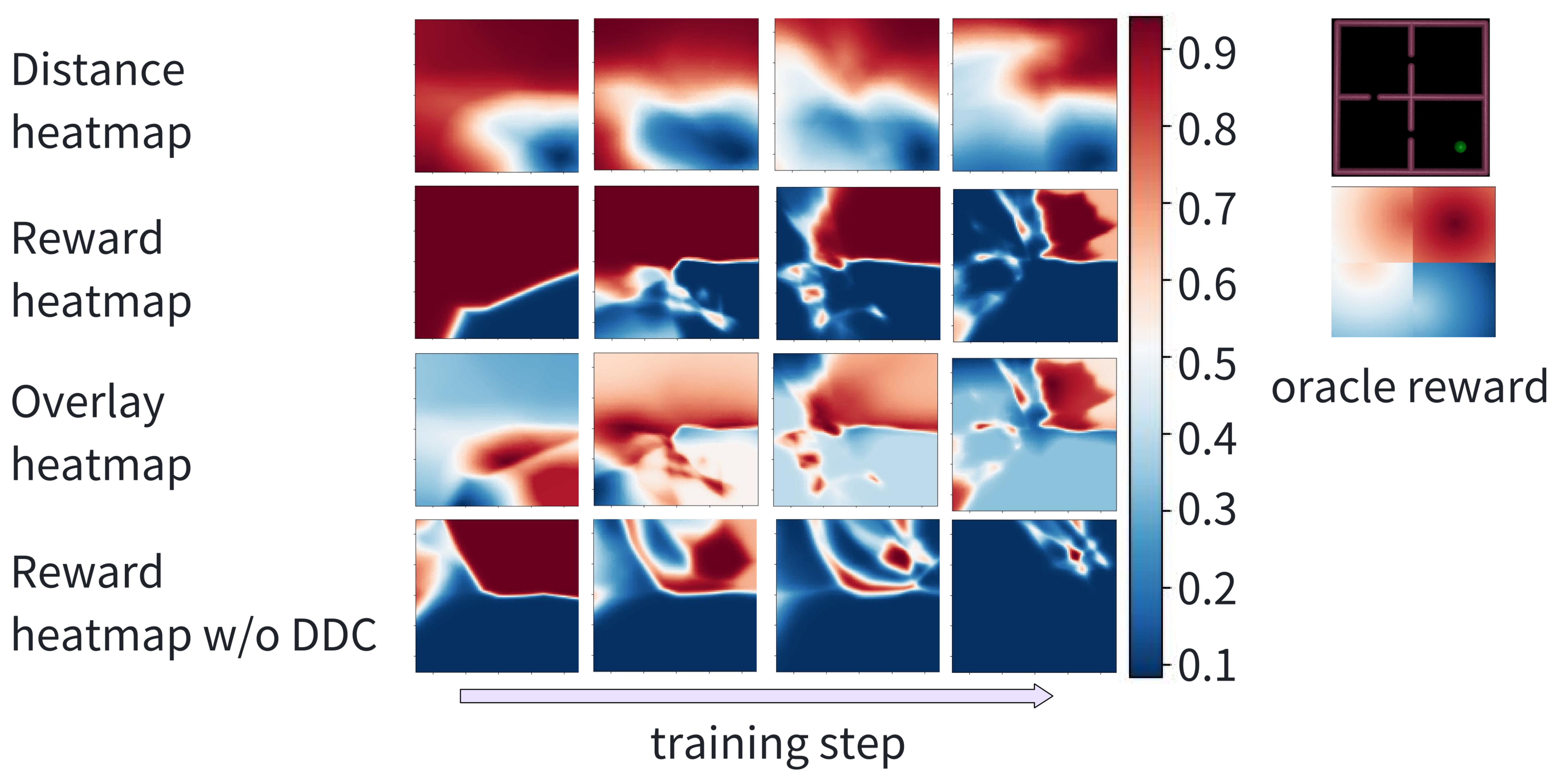}
    \caption{\textcolor{black}{DDC effects on reward model in the Four rooms domain. The heatmaps serve as visual representations of the models' learning progress, where the distance model learns to accurately estimate the state to initial state distances and the reward model learns to assign values that incentivize the agent's progression towards the final goal. The overlay reward heatmap captures the integrated effect of both the reward model and the distance model, as articulated in the adjustments of the high-level policy update detailed in \Cref{adjust constraint high policy update}.}}
    \label{hotward}
\end{figure}

We also analyze the heatmap of the reward and distance model. The analysis of the heatmaps in \Cref{hotward} shows complex interactions between the distance and reward models in the {\M} framework. The distance model assesses subgoal difficulty. This overlap, calculated with $r_{\hf}(s, g^{\sub},g) + \alpha H(g^{\sub},g,k)$, identifies areas optimizing rewards in alignment with the low-level policy's capabilities. However, without DDC, the human feedback-based reward model is limited, indicating only high-reward areas without guiding the agent on how to reach them. This can also corroborate the subgoal distribution in \Cref{backgroundddc}.

\subsection{Impact of Human Feedback (RQ3)}
This part is dedicated to examining the effects of human feedback on Model {\M}. Firstly, we evaluate the influence of the frequency and quantity of human feedback on algorithmic performance. Following this, we examine the differences between real human labels and synthetic labels, explain the rationale for using synthetic labels in our experiments, and demonstrate the practical applicability of our algorithm to real-world contexts.

\begin{table}[tp!]
\centering
\caption{Cross-analysis on episodes (100 steps per episode) to the success of variables batch queries and query frequency. For each combination of variables, we ran 5 trials on different seeds and recorded the mean and variance.}
\label{tab:cross-analysis}
\begin{tabular}{c|ccc}
\hline
& \multicolumn{3}{c}{Query Frequency} \\
{Batch Queries} & \textbf{25} & \textbf{50} & \textbf{100} \\
\hline
\textbf{0} & 3522$\pm$1441& - & -\\
\textbf{10} & 1565$\pm$128 & 1765$\pm$268 & 1820$\pm$261\\
\textbf{25} & 1455$\pm$165 & 1770$\pm$108 & 1683$\pm$203\\
\textbf{50} & \textbf{1406}$\pm$\textbf{55} & 1735$\pm$253 & 1760$\pm$277\\
\hline
\end{tabular}
\end{table}

\textbf{Analyzing the quantity and frequency of feedback}. In \Cref{tab:cross-analysis}, the data shows the number of training episodes needed for 100$\%$ task success across different query frequencies and batch sizes. The table illustrates that integrating human feedback into training markedly improves learning speed, as highlighted by the contrast between label-free experiments and those with feedback. Further analysis demonstrates a consistent pattern: higher feedback frequency and quantity correlate with increased success rates. When considering the total labels and their impact on learning speed, our algorithm significantly enhances efficiency and stability by requiring only a small number of labels: 10 per 100 episodes, totaling 180, compared to experiments with non-feedback.

\begin{figure}[tp!]
    \centering
    \begin{minipage}[t]{0.23\textwidth}
        \centering
        \includegraphics[width=\linewidth]{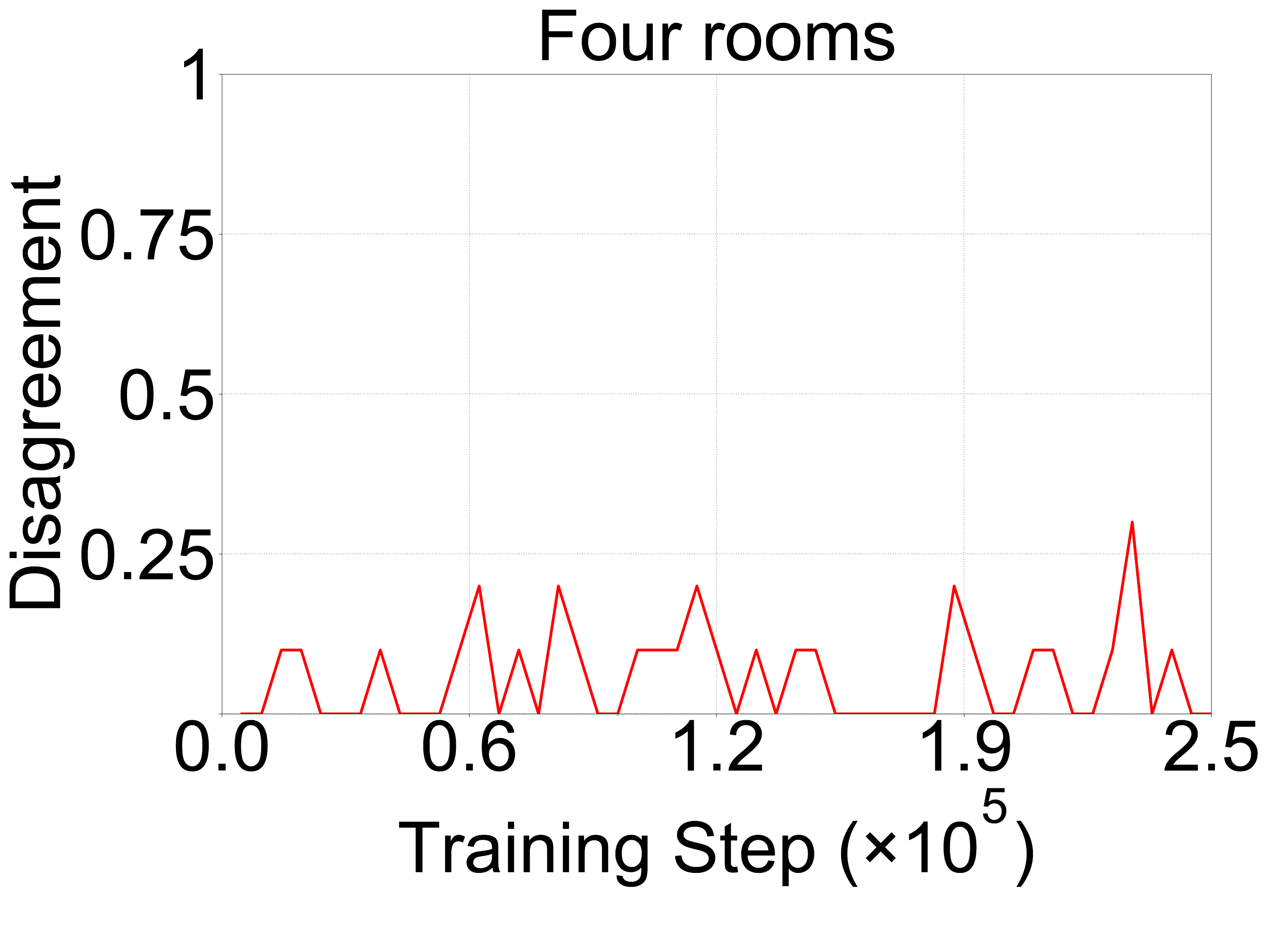}
    \end{minipage}
    \begin{minipage}[t]{0.23\textwidth}
        \centering
        \includegraphics[width=\linewidth]{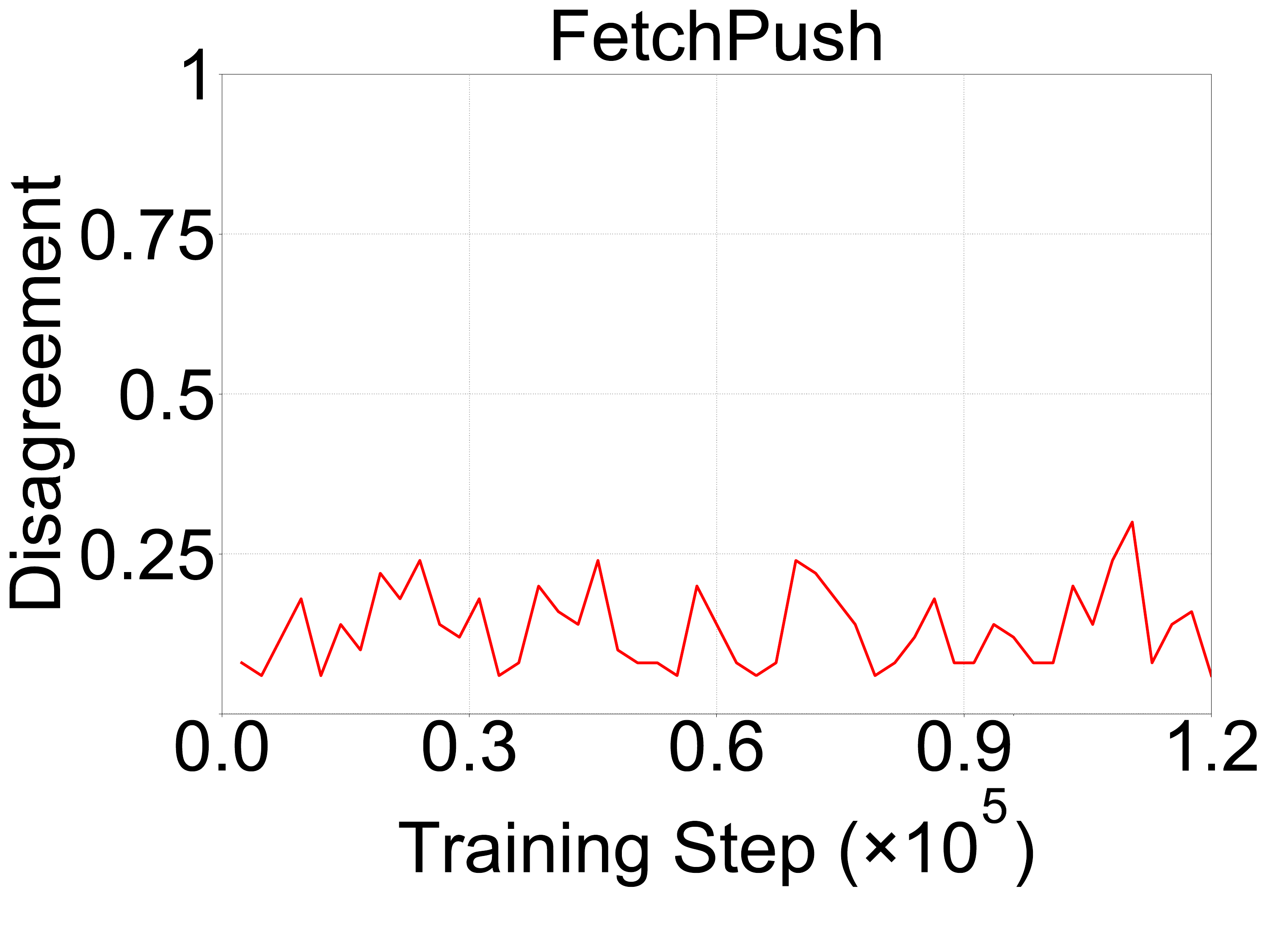}
    \end{minipage}
    \hfill 
    \begin{minipage}[t]{0.23\textwidth}
        \centering
        \includegraphics[width=\linewidth]{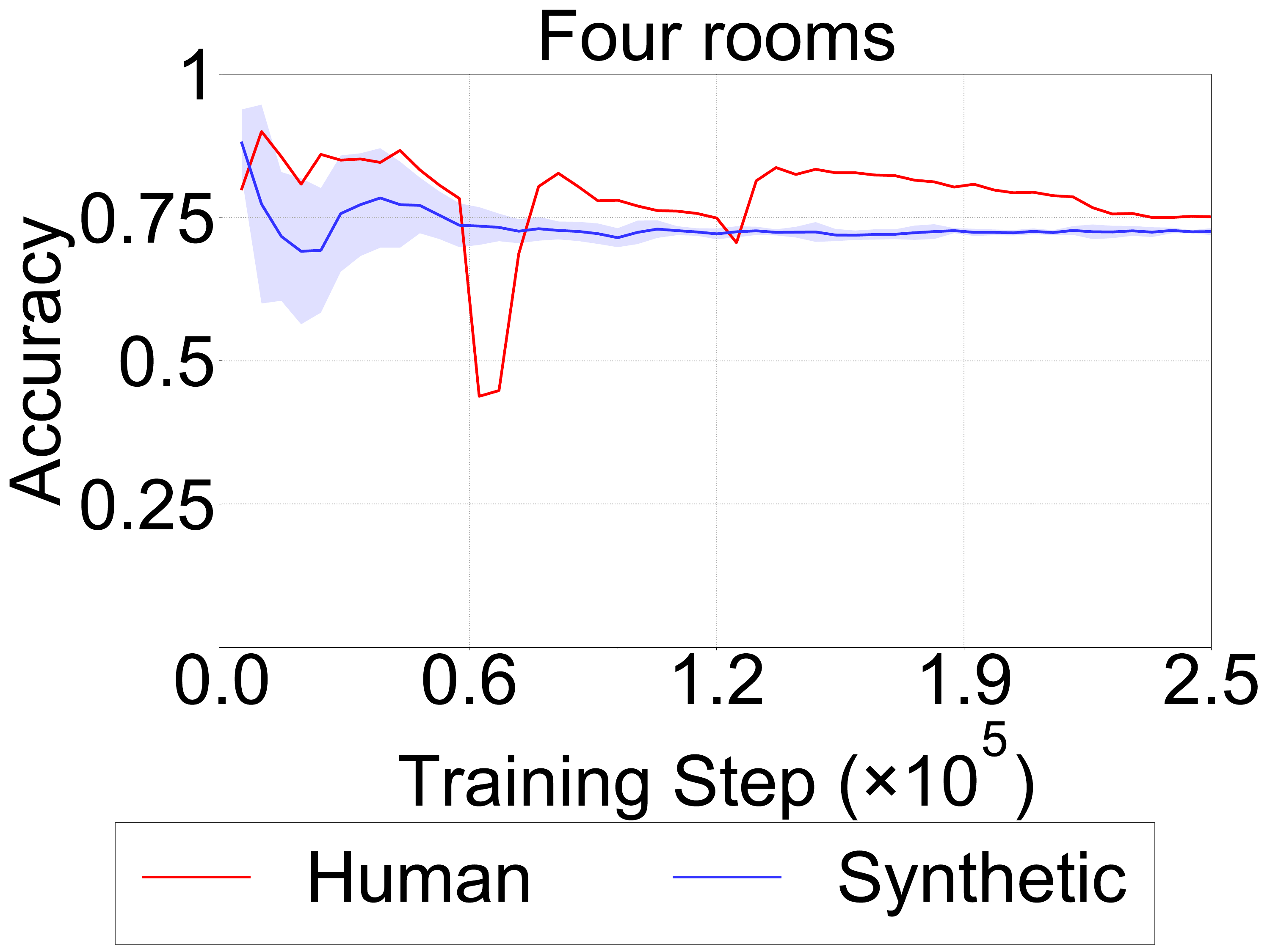}
    \end{minipage}
    \begin{minipage}[t]{0.23\textwidth}
        \centering
        \includegraphics[width=\linewidth]{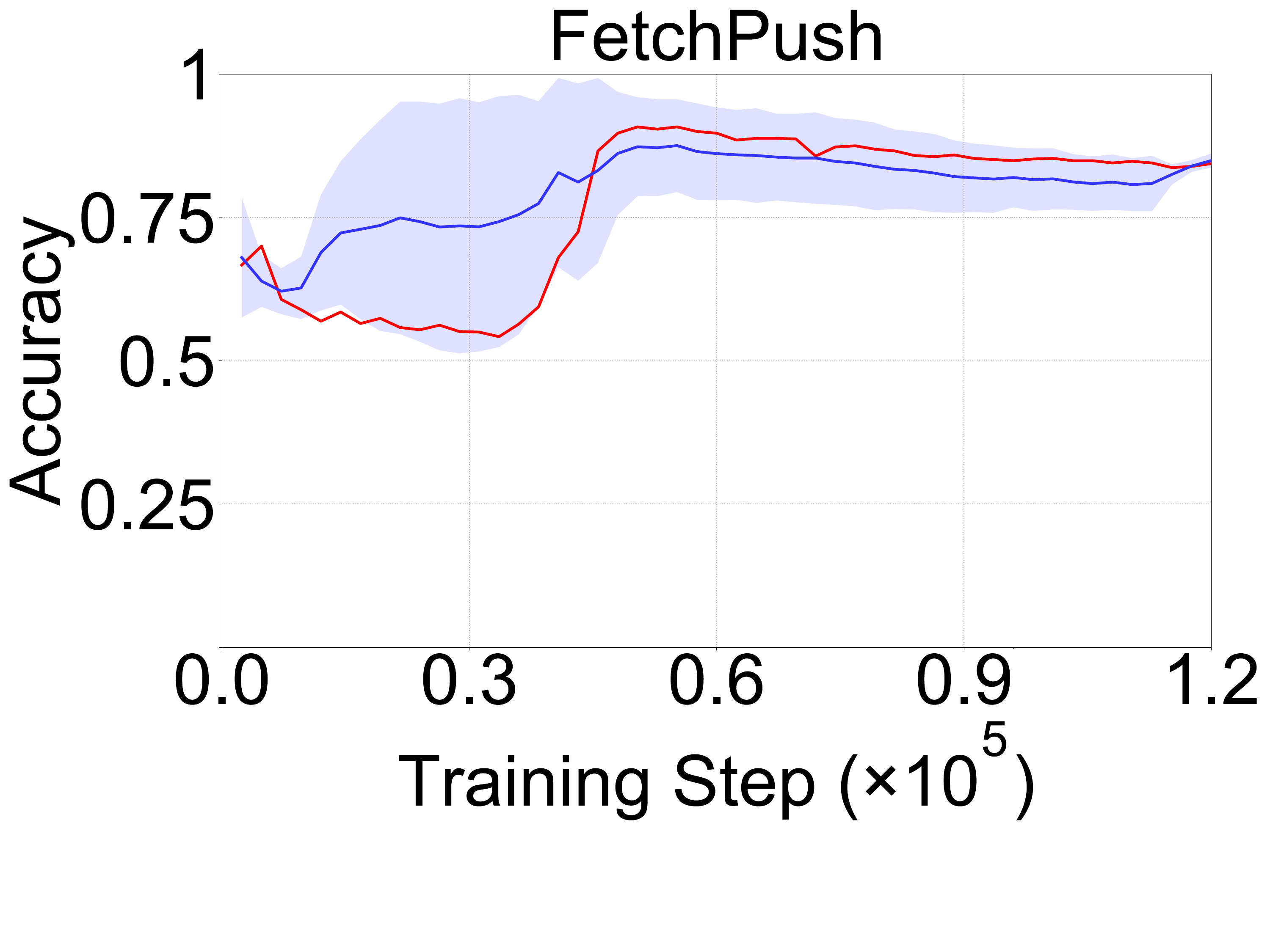}
    \end{minipage}
    \caption{On the top side, it records the rate of disagreement between human annotations and synthetic labels. There lies a 10\% to 20\% disparity between human-labeled labels and synthetic labels. On the bottom side, we present the accuracy rate on the training data of the reward model. Within the Four room domain, the reward model's accuracy is 75\%, and approximately 80\% for FetchPush.}
    \label{disagreement and accuracy}
\end{figure}

\textbf{Comparing human collected labels and synthetic labels}. In our study, we employ two methods for providing human guidance: synthetic labels generated through scripted techniques and human-generated labels. 

We monitored the disagreement rates between authentic human labels and synthetic labels, as well as the accuracy of the reward models, as shown in \Cref{disagreement and accuracy}. Given that the starting and ending positions in the Four rooms domain are predetermined, we provided 10 labels every 25 episodes. Conversely, in FetchPush where positions are not fixed, we provided 50 labels every 50 episodes. The data reveals discrepancies between human and synthetic labels, suggesting that human labeling is susceptible to errors due to subjective factors. Furthermore, the RLHF method produced a model whose accuracy converged to approximately 80\%, indicating that even the trained model has its uncertainties and potential inaccuracies. This is also supported by the difference between the heatmap of the trained reward model and the heat map of the oracle reward model in \Cref{hotward}. However, our algorithm demonstrates sufficient robustness to handle the inaccuracies inherent in RLHF. This implies that integrating RLHF into the high-level policy is a highly compatible approach; on the one hand, RLHF fulfills the reward requirements of the high-level, and on the other hand, the low-level policy remains unaffected by the inaccuracies of RLHF. During the algorithm testing, the low-level policy completes tasks independently, hence the final performance of the algorithm is not affected.

\begin{figure}[tp!]
    \centering
    \centering
    \begin{minipage}[t]{0.3\textwidth}
        \centering
        \includegraphics[width=\linewidth]{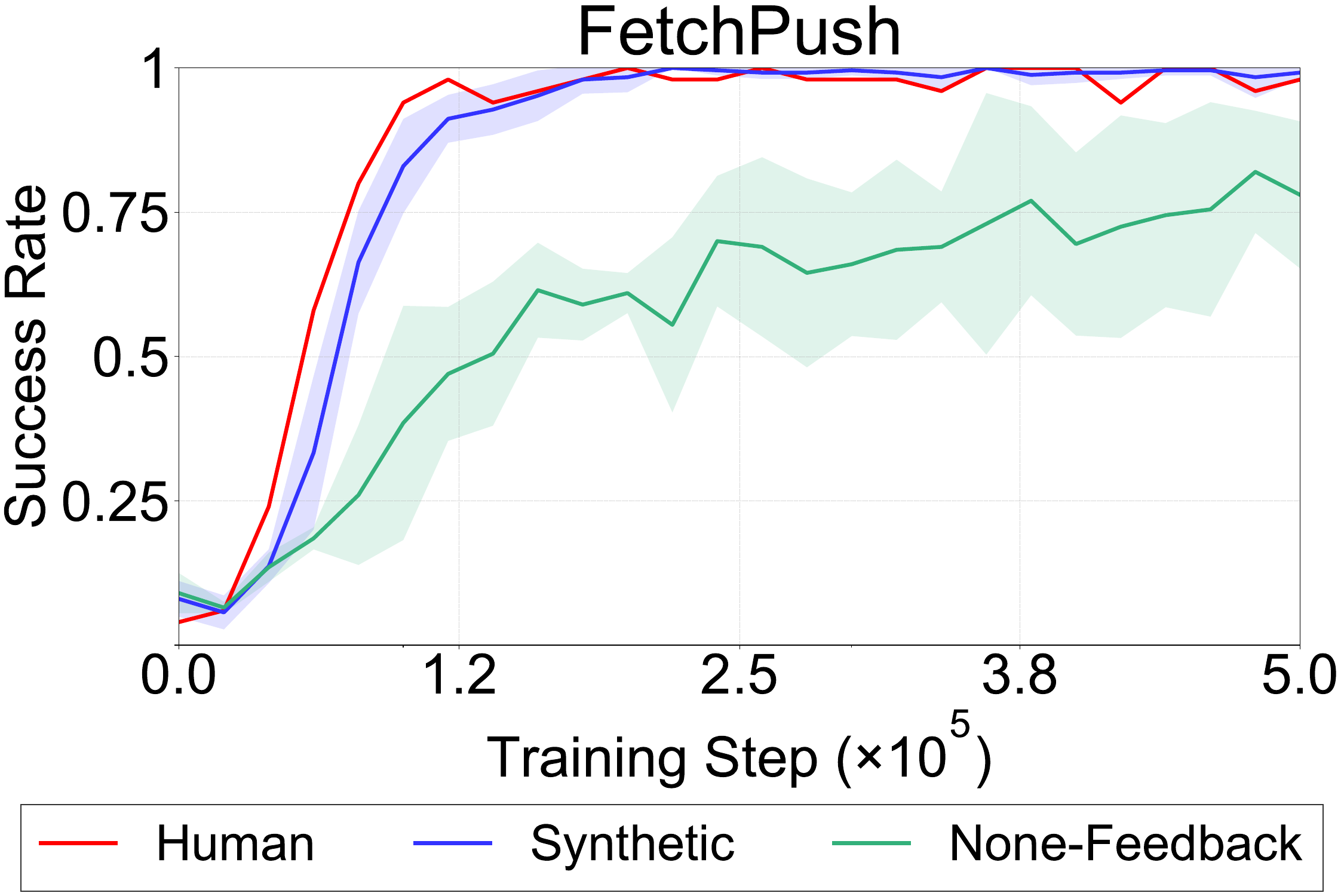}
    \end{minipage}
    \begin{minipage}[t]{0.3\textwidth}
        \centering
        \includegraphics[width=\linewidth]{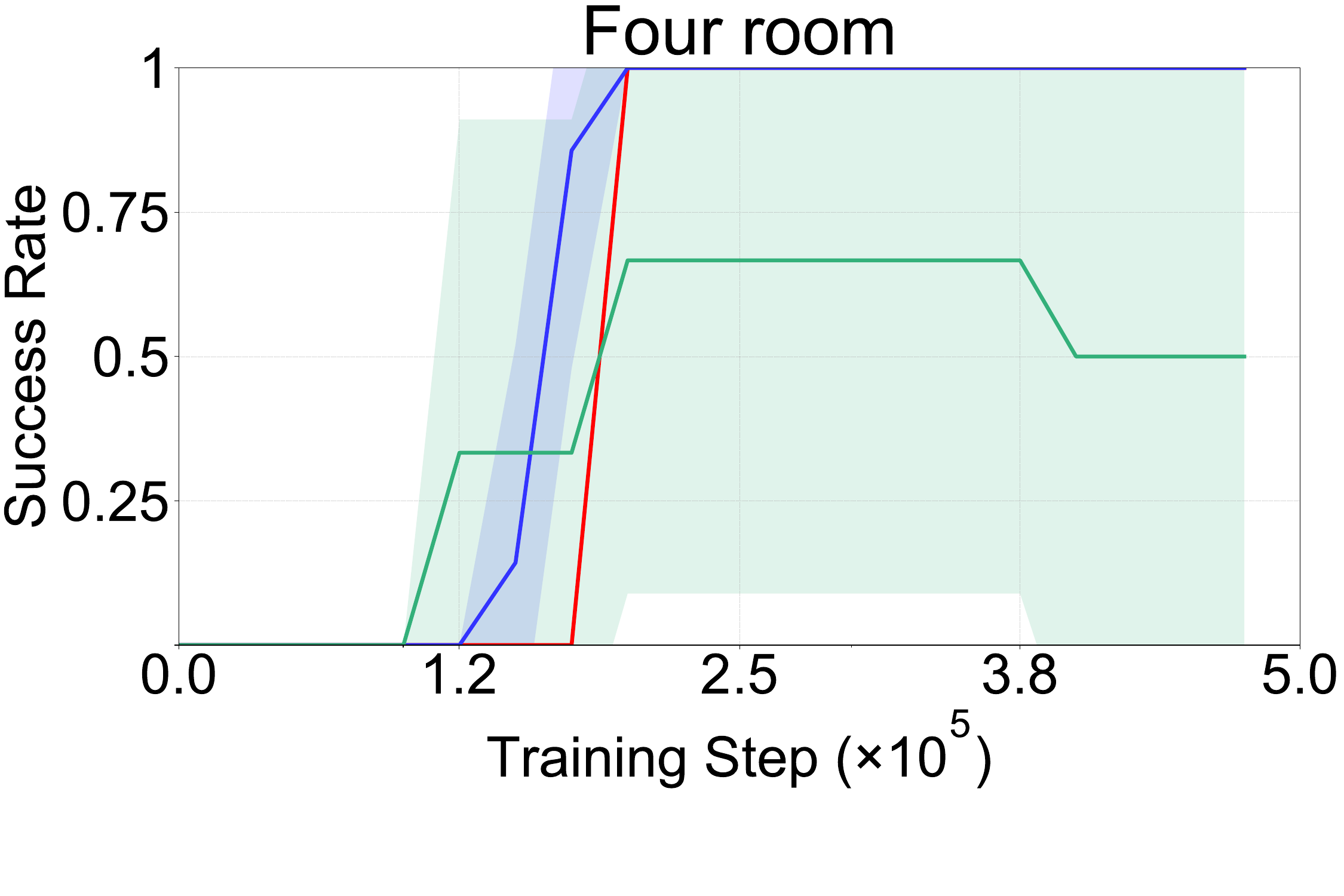}
    \end{minipage}
    \caption{Experiments for evaluating {\M} with script-generated and human-collected labels on the FetchPush and Four rooms domain. In these experiments, we provided 10 labels every 25 episodes. The results for the script-generated labels and non-feedback scenarios were based on five random seeds to ensure robustness, while the human-collected label experiment relied on a single random seed.}
    \label{exp4}
\end{figure}

We compare the performance of human collected labels and synthetic labels in FetchPush and Four rooms domains. As illustrated in \Cref{exp4}, models trained with human labels exhibited comparable learning rates, with performance differences potentially attributed to noise in human feedback. After comparing the experiments with human collected labels to the experiments with non-feedback, we can find a significant improvement in the algorithm's convergence speed and final performance. Therefore, we conclude that synthetic labels have a similar effect on improving the algorithm as real human labels. This implies that in other experiments, synthetic labels could potentially serve as a replacement for human collected labels.

\subsection{Ablation Studies (RQ4)}
\begin{figure}[!htbp]
    \centering
    \begin{minipage}[t]{0.23\textwidth}
        \centering
        \includegraphics[width=\linewidth]{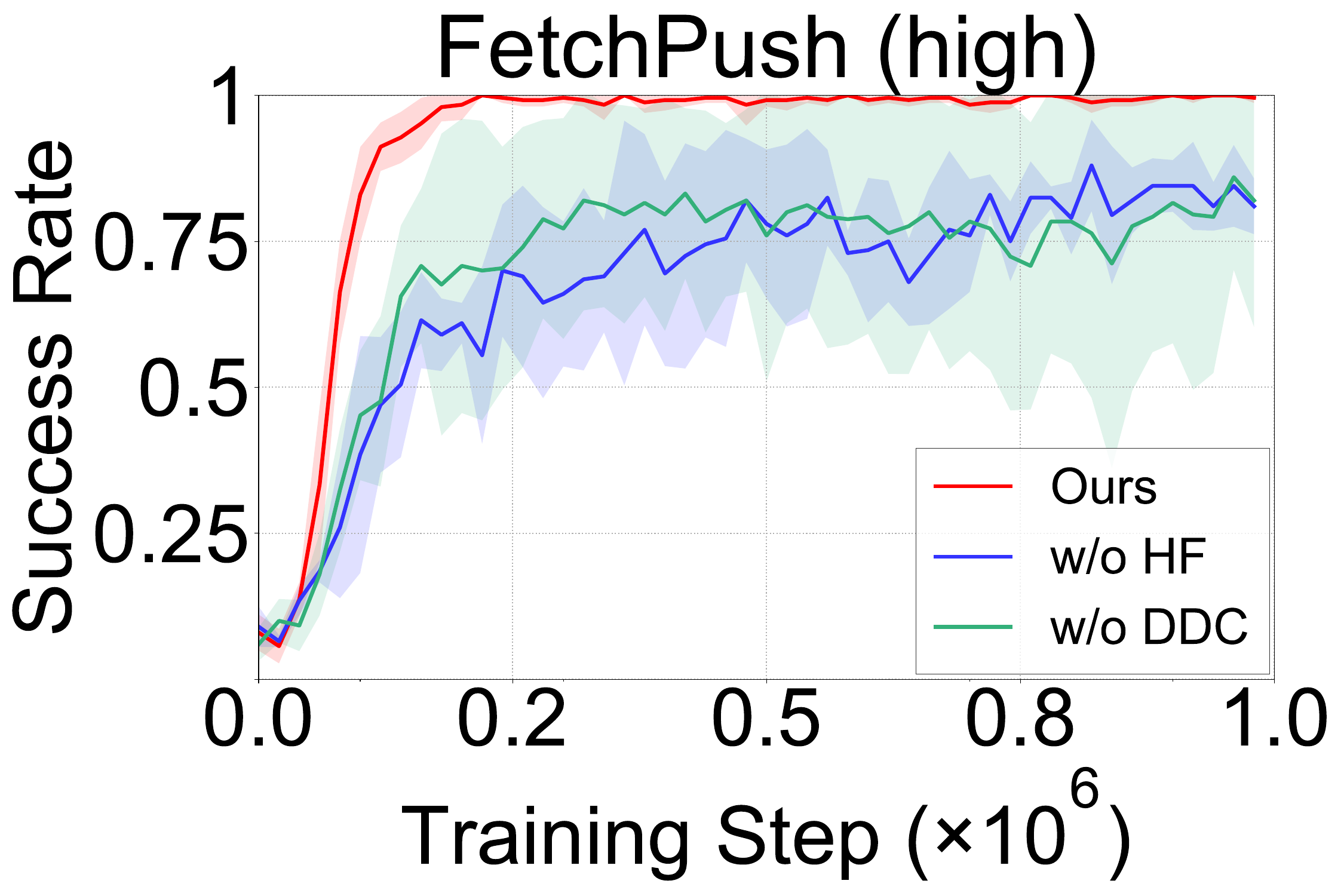}
    \end{minipage}
    \hfill 
    \begin{minipage}[t]{0.23\textwidth}
        \centering
        \includegraphics[width=\linewidth]{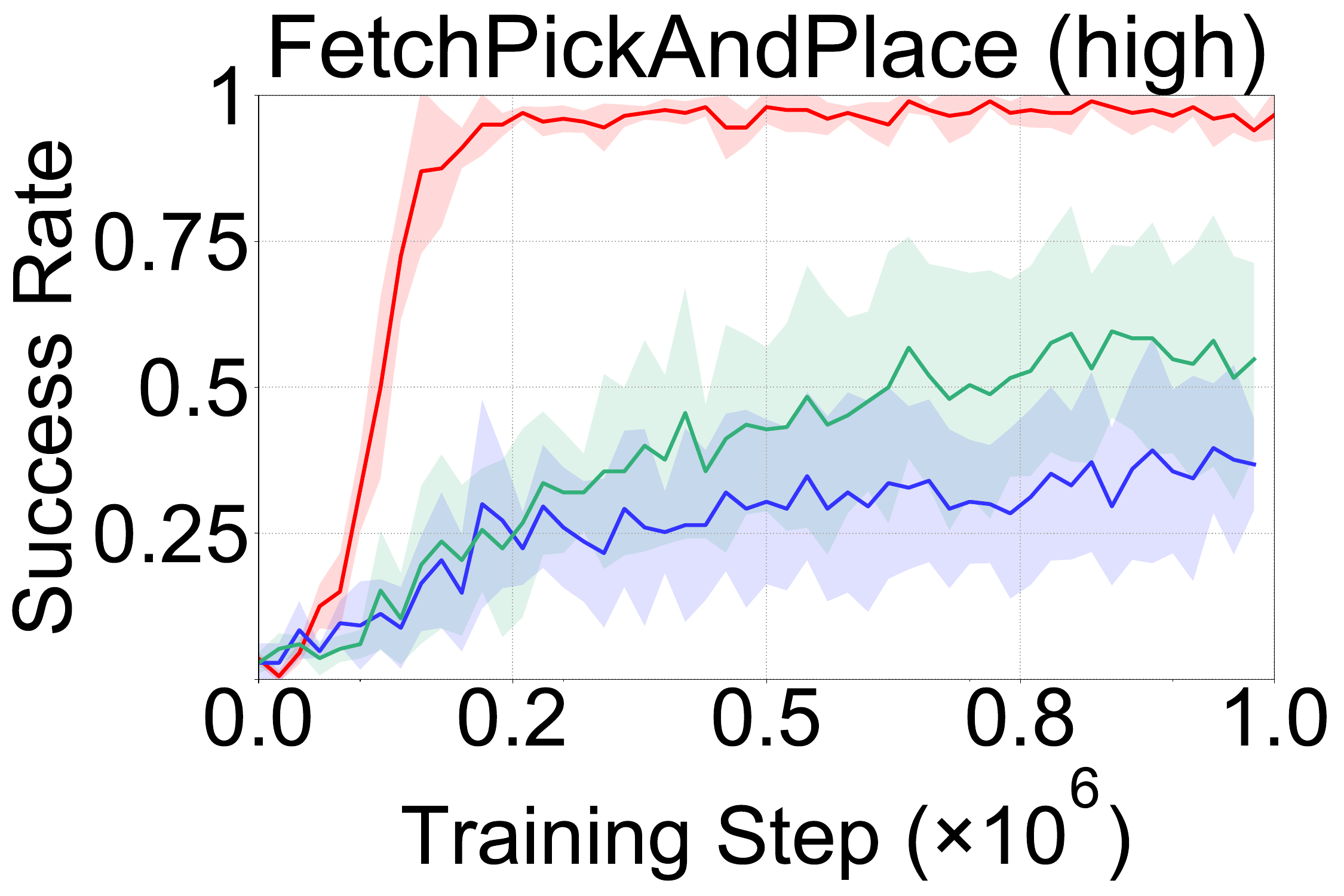}
    \end{minipage}
    \begin{minipage}[t]{0.23\textwidth}
        \centering
        \includegraphics[width=\linewidth]{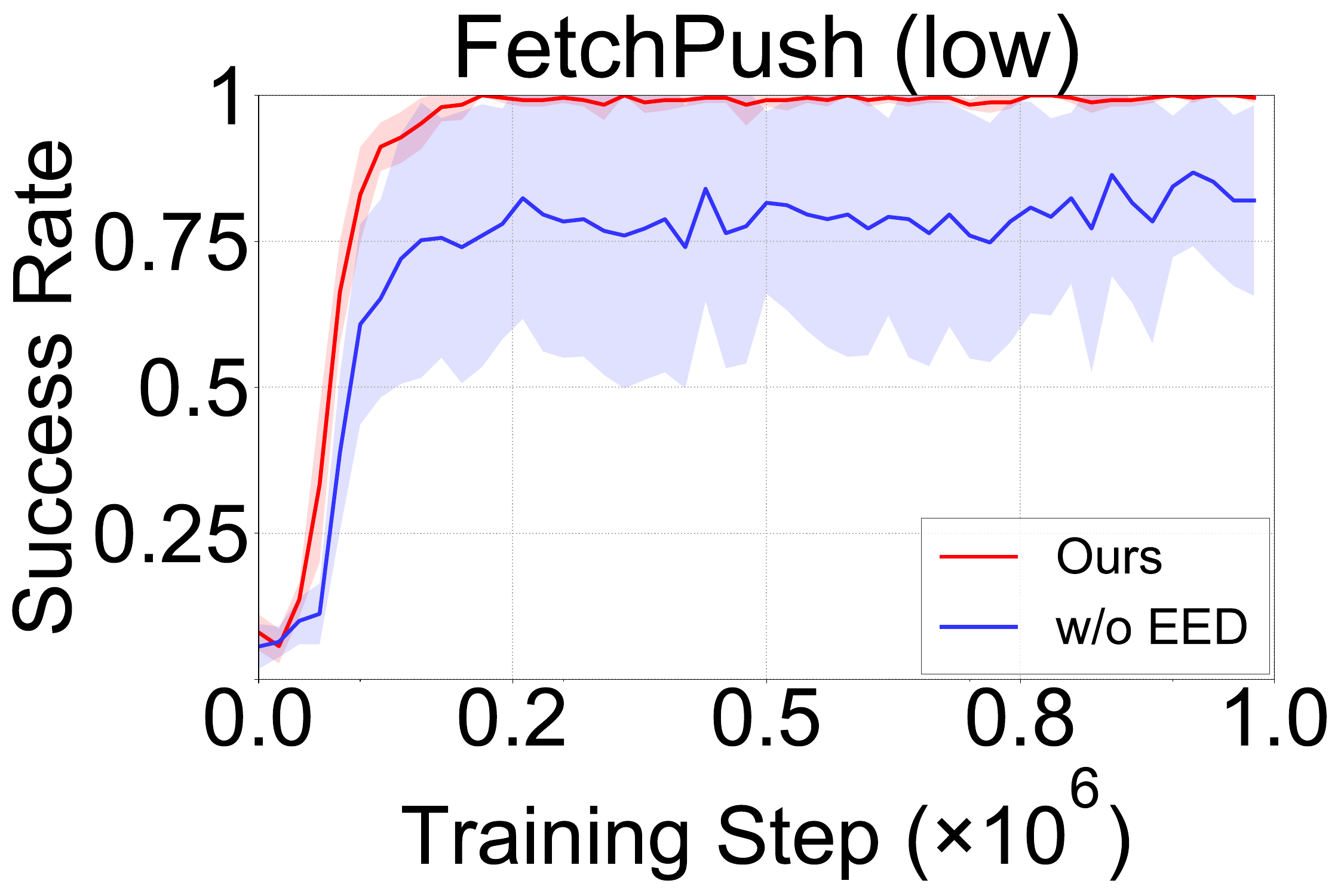}
    \end{minipage}
    \hfill 
    \begin{minipage}[t]{0.23\textwidth}
        \centering
        \includegraphics[width=\linewidth]{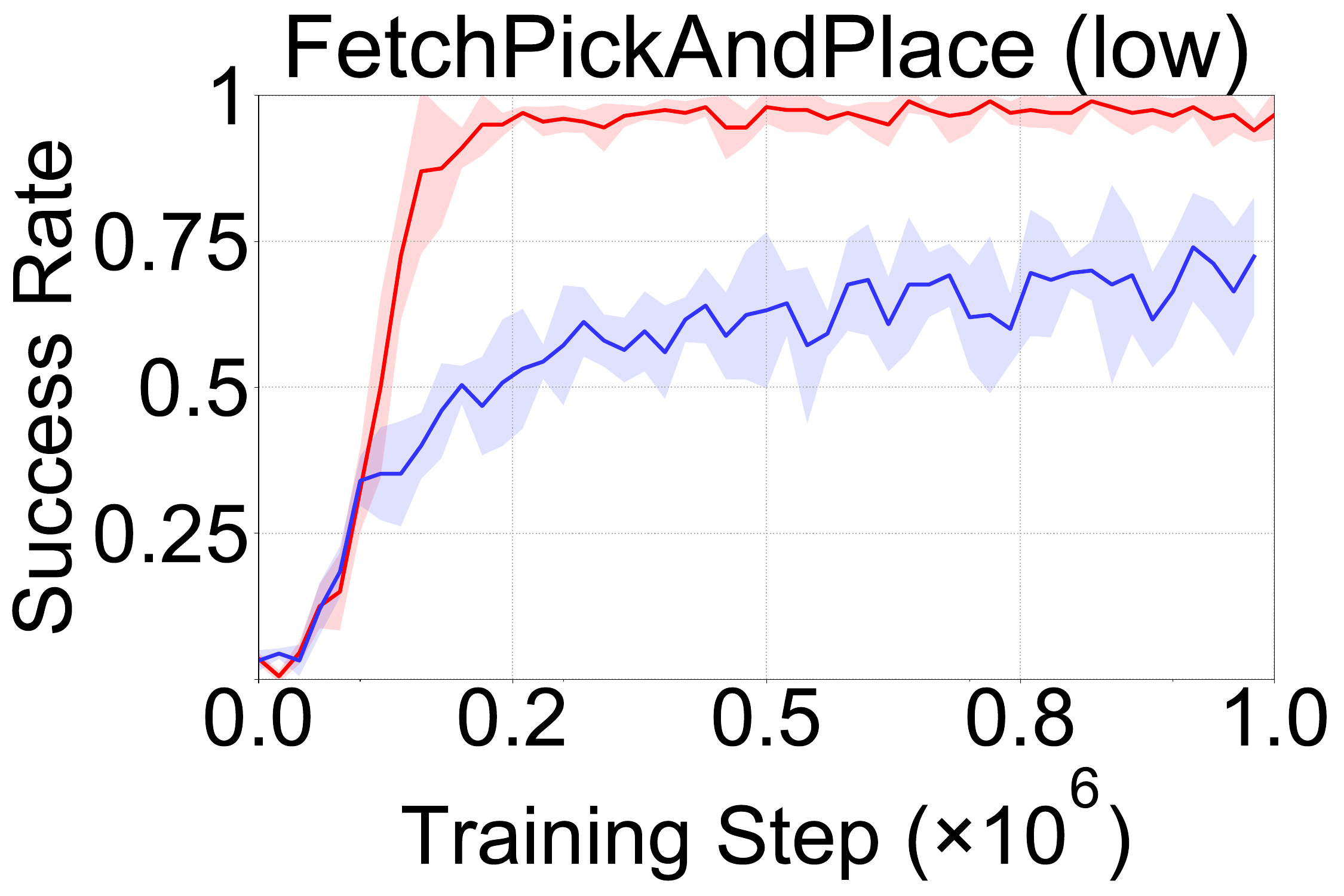}
    \end{minipage}
    \caption{Ablations studies for {\M} in FetchPickAndPlace and FetchPush domains at five random seeds. The ablation experiment involved immobilizing the low-level policy, removing the HF and DDC functions of the high-level policy, immobilizing the high-level policy, and removing the EED at the low-level.}
    \label{exp5}
\end{figure}

The ablation study in the {\M} framework, focusing on the FetchPickAndPlace and FetchPush domains, evaluates how components like HF (human feedback), DDC, and EED ({\EED}) contribute to learning efficiency and goal achievement. \Cref{exp5} (top) assesses each module's effectiveness by systematically removing high-level modules and observing the impact on model performance. The removal of DDC and human feedback results in slower convergence, reduced performance, and decreased stability. The study mentioned in \Cref{exp5} (bottom) discovered that the absence of EED negatively affects the balance between exploration and exploitation, resulting in lower success rates even after algorithm convergence. This emphasizes the significance and interdependence of these modules in improving learning for complex tasks in the {\M} framework.

\section{Conclusion}
This study presents \M, an innovative method that combines human feedback and Dynamic Distance Constraint for learning guidance. It integrates high-level human insights for selecting subgoals concerning low-level capabilities and introduces exploration-exploitation decoupling at the low-level to improve training stability and efficiency. Our experiments demonstrate the framework's effectiveness in complex tasks with sparse rewards, outperforming existing baselines. 

We recognize the complexity of human guidance beyond just subgoal selection and aim to explore a wider range of feedback integration to enhance learning dynamics. We plan to further expand the framework's applications and refine its mechanisms, aiming to advance hierarchical reinforcement learning and create more intuitive, adaptable learning systems.

 
\appendices

\section{Domains Details}
\label{appendix:B}
In this section, we will provide further elaboration on the benchmarks utilized for comparing our method with the baselines. Specifically, we will delve into the details of the observation space, action space, and the configuration of the reward function.

\textbf{FetchPush}. In this domain, the aim is to use a 7-DoF Fetch Mobile Manipulator, equipped with a closed two-fingered parallel gripper, for transporting a block to a set position on a table. The robot's movement is finely adjusted using Cartesian coordinates, and the MuJoCo framework calculates the inverse kinematics. This task, which involves pushing the block with the gripper constantly closed, is ongoing, requiring the robot to steadily keep the block at the target position. The scenario is observed through a 25-element array, encompassing kinematic details of both the block and gripper. The action space is a Box(-1.0, 1.0, 4, float32), with each action altering the gripper's Cartesian position $(dx, dy, dz)$ and managing its opening and closing. The reward system applies -1 when the block isn't at the target, and 0 when correctly positioned, defined as being within 0.05 meters of the target.

\textbf{FetchPickAndPlace}. Utilizing the same robot setup with FetchPush, this domain focuses on moving a block to a defined point, including mid-air locations. It shares the same observation array and action space as FetchPush, with the addition of a goal-aware observation space. The reward system remains consistent with the FetchPush domain.

\textbf{FetchDraw}. Utilizing the same 7-DoF Fetch Mobile Manipulator with a two-fingered parallel gripper, this task involves the robot's precise interaction with a drawer. The objective is two-fold: (1) to reach for the drawer handle, and (2) to slide the drawer to a target position by pulling or pushing the handle. The manipulation requires the gripper to perform open and close actions for a firm grasp on the drawer handle. The robot must successfully move and maintain the drawer at the target position indefinitely. The reward system remains consistent with the FetchPush domain.

\textbf{FetchObsPush}. This task engages the same 7-DoF Fetch Mobile Manipulator, which is equipped with a two-fingered parallel gripper. The robot is tasked with manipulating a block to a desired position on a table in the presence of an obstacle. The challenge requires the robot to (1) approach and securely grasp the block, and (2) navigate and push the block to the target location, accounting for the obstacle whose size and shape are unknown. This task demands precision handling and adaptability to avoid obstacle while ensuring the block reaches its target position. As with the FetchPush, FetchObsPush is continuous, with the robot required to keep the block within 0.05 meters of the target position indefinitely. The reward system remains consistent with the FetchPush domain.

\textbf{Pusher}. This domain involves the manipulation of a robotic arm, specifically a sawyer robotic arm, in an environment with multiple obstacles. The objective is to successfully push an obstacle, referred to as a puck, to a designated goal area marked by a red dot. Unlike the FetchPush problem, which has a 25-dimensional observation, the state space in this environment is determined by the position of the puck and the arm. The action space, on the other hand, involves controlling the position of the robotic arm. It is a discrete 9-dimensional space where each action corresponds to a delta change in the position in a 2-D space. The reward system remains consistent with the FetchPush domain.

\textbf{Four rooms}. In the four-room domain, the agent's task is to navigate to a specified goal while dealing with initially unknown obstacles. The agent is positioned at (0.4, -0.4) in the bottom right room, aiming for the goal at (0.25, 0.25) in the top right room. The state observation is the agent's precise (x, y) location, and it has a set of 9 possible actions to choose from, which allow movement in all cardinal and diagonal directions, or to remain stationary. Key to this task are the doorways in the walls, which are the sole means for the agent to traverse between rooms.

In the FetchPush, FetchPickAndPlace, FetchDraw, FetchObsPush, and Pusher domains, synthetic labels are generated using a dense sparse approach. This means that the reward returned is calculated as the negative Euclidean distance between the achieved goal position and the desired goal. In the Four rooms domain, we utilize the reward function displayed in \Cref{four room reward}. The reward heatmap, referred to as the oracle reward model, is depicted in \Cref{hotward}. In the top right quadrant, the agent receives a reward based on the negative Euclidean distance from its current position $ s $ to the goal $ [0.25, 0.25] $. In the top left quadrant, the reward is the negative Euclidean distance from $ s $ to a fixed point $ [0, 0.3] $, with an additional penalty of -0.3. In the bottom left quadrant, the reward is the negative Euclidean distance from $ s $ to $ [-0.3, 0] $, with a penalty of -0.6. In the bottom right quadrant, it is the negative Euclidean distance from $ s $ to $ [0, -0.3] $, with a penalty of -1.

\begin{equation}
\begin{aligned}
    &\text{r}_\text{Four rooms}(s) = 
    \\
    &\begin{cases}
    -\lVert s - [0.25, 0.25] \rVert_2 & \text{if } s_x \geq 0 \text{ and } s_y \geq 0, \\
    -\lVert s - [0, 0.3] \rVert_2 - 0.3 & \text{if } s_x \leq 0 \text{ and } s_y \geq 0, \\
    -\lVert s - [-0.3, 0] \rVert_2 - 0.6 & \text{if } s_x \leq 0 \text{ and } s_y \leq 0, \\
    -\lVert s - [0, -0.3] \rVert_2 - 1 & \text{if } s_x \geq 0 \text{ and } s_y \leq 0.
    \end{cases}
\end{aligned}
\label{four room reward}
\end{equation}

\section{Details of Implementation}
\paragraph{Labels acquisition}\label{RLHF details}
We obtain labels through two ways: human annotation and script synthesis. The script-synthesized labels are generated based on dense reward functions. 

After extracting tuple pairs from the replay buffer $B_{\hi}$ in line 22 of \Cref{alg:A}, such as ($s_{\hi_{1}}, g^{\sub}_1, g_1$ and $s_{\hi_{2}}, g^{\sub}_2, g_2$), the subsequent step involves obtaining preference labels through interaction with humans or a script. We have designed an interactive interface for human annotation, see \Cref{appendix interaction}. This interface is constructed based on the Four rooms domain, where it differentiates tuple pair information through distinct colors of hollow and solid circles. Since the goal position is fixed within this domain, we represent the goal located in the top right corner with a red solid circle. When an annotator needs to determine which tuple is more advantageous for policy learning, they can either click the gray button situated to the right of the letters "goal" or input '0' or '1' on the keyboard. If the annotator cannot distinguish which tuple is good or bad, he can enter the '2' on the keyboard, and we will set the label to 0.5 to indicate the same preference.

We employ a script to fulfill the role of annotators. The script will calculate the dense reward using an oracle reward function. In the first five domains, we employ the Euclidean distance between the subgoal and the goal. In the Four rooms domain, we use the reward function as mentioned in \Cref{four room reward}. Depending on the magnitude of the calculated reward for different tuples, we can assign the label to the tuple corresponding to the higher reward.

\begin{figure}[tp!]
    \centering
    \includegraphics[width=0.5\linewidth]{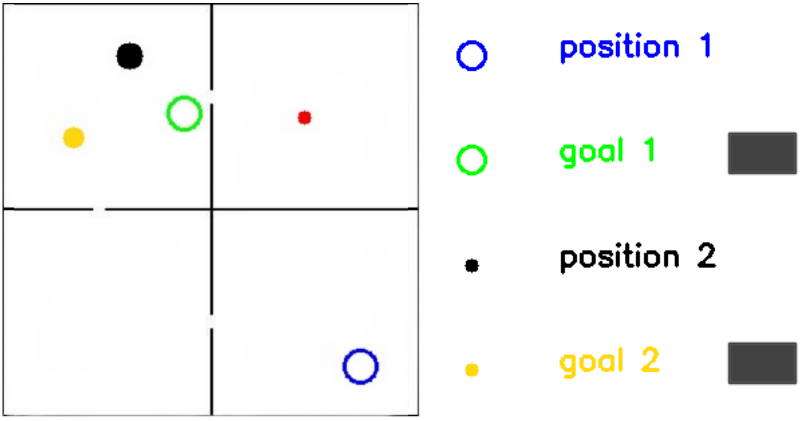}
    \caption{A simple interface for human annotation.}
    \label{appendix interaction}
\end{figure}

\begin{table}[tp!]
\caption{Hyperparameters of {\M}.}\label{tab:hyperparameters}
\centering
\begin{tabular}{@{}ll@{}}
\hline
Hyperparameters             & Values\\ 
\hline
\hline
\textbf{High-level policy}      &        \\
\hline
Actor learning rate         & $3 \times 10^{-4}$  \\
Critic learning rate        & $3 \times 10^{-4}$ \\
Replay buffer size          & $10^{6}$         \\
Hidden layers                 & 3          \\
Hidden size                 & 256          \\
Batch size                  & 256              \\
Soft update rate            & 0.005             \\
Policy update frequency     & 1                 \\
$\gamma$                    & 0.95              \\
Distance model learning rate& $3 \times 10^{-4}$ \\
Distance model replay buffer size& 1000 \\
Distance model hidden layers  & 3          \\
Distance model hidden size  &  256 \\
Distance model batch size  &  256 \\
Reward model learning rate& $3 \times 10^{-4}$ \\
Reward model replay buffer size& 1000 \\
Reward model hidden layers  & 3          \\
Reward model hidden size  &  256 \\
Reward model batch size  &  256 \\
Query Frequency & 50 \\
Batch Queries & 50 \\
Success rate high threshold & 0.6 \\
Success rate low threshold & 0.3 \\
$\Delta k$ & 0.05 \\
Initial $k$ & 0.05 \\
$\beta$ & 0.1 \\
\hline
\hline
\hline
\textbf{Low-level policy}   &                   \\
\hline
Actor learning rate         & $10^{-3}$          \\
Critic learning rate        & $10^{-3}$            \\
Hidden layers               & 3          \\
Hidden size                 & 256          \\
Replay buffer size          & $10^{6}$           \\
Batch size                  & 512               \\
Soft update rate            & 0.005             \\
Policy update frequency     & 1                 \\
$\gamma$                    & 0.95              \\
RND learning rate           & $3 \times 10^{-4}$ \\
RND hidden layers           & 3              \\
RND hidden size             & 256              \\
RND represent size          & 512              \\
RND bonus scaling           & 1.0               \\
Hindsight sample ratio      & 0.8               \\
\hline
\end{tabular}
\end{table}

\paragraph{Hyperparameters} see Table \ref{tab:hyperparameters}.

\bibliographystyle{IEEEtran}
\bibliography{reference}


\end{document}